\documentclass{article}

\usepackage{microtype}
\usepackage{graphicx}
\usepackage{subfigure}
\usepackage{booktabs} 

\usepackage{hyperref}

\hypersetup{
colorlinks   = true, 
urlcolor     = blue, 
linkcolor    = blue, 
citecolor   = blue 
}

\usepackage[accepted]{icml2024}

\usepackage{amssymb}
\usepackage{amsmath}
\usepackage{amsthm}
\usepackage{amsfonts}
\usepackage{bm}

\usepackage{makecell}
\usepackage{float}
\usepackage{rotating}

\DeclareMathOperator*{\argmax}{\text{argmax}}
\DeclareMathOperator*{\argmin}{\text{argmin}}

\newcommand{\ourmethod}{\texttt{MATCH-OPT}}
\usepackage[capitalize,noabbrev]{cleveref}

\theoremstyle{plain}
\newtheorem{theorem}{Theorem}[section]

\newtheorem{lemma}[theorem]{Lemma}

\theoremstyle{definition}
\newtheorem{definition}[theorem]{Definition}
\newtheorem{assumption}[theorem]{Assumption}
\theoremstyle{remark}

\usepackage[textsize=tiny]{todonotes}

\icmltitlerunning{Learning Surrogates for Offline Black-Box Optimization via Gradient Matching}

\begin{document}

\twocolumn[
\icmltitle{Learning Surrogates for Offline Black-Box Optimization via Gradient Matching}




\begin{icmlauthorlist}
\icmlauthor{Minh Hoang}{princeton}
\icmlauthor{Azza Fadhel}{wsu}
\icmlauthor{Aryan Deshwal}{wsu}
\icmlauthor{Janardhan Rao Doppa}{wsu}
\icmlauthor{Trong Nghia Hoang}{wsu}
\end{icmlauthorlist}


\icmlaffiliation{princeton}{Lewis-Sigler Institute of Integrative Genomics, Princeton University, New Jersey, USA}
\icmlaffiliation{wsu}{School of Electrical Engineering and Computer Science, Washington State University, Pullman, Washington, USA}

\icmlcorrespondingauthor{Trong Nghia Hoang}{trongnghia.hoang@wsu.edu}
\icmlcorrespondingauthor{Janardhan Rao Doppa}{jana.doppa@wsu.edu}

\icmlkeywords{Black-Box Optimization}

\vskip 0.3in
]



\printAffiliationsAndNotice{}  

\begin{abstract}
Offline design optimization problem arises in numerous science and engineering applications including material and chemical design, where expensive online experimentation necessitates the use of \textit{in silico} surrogate functions to predict and maximize the target objective over candidate designs. Although these surrogates can be learned from offline data, their predictions are often inaccurate outside the offline data regime. This challenge raises a fundamental question about the impact of imperfect surrogate model on the performance gap between its optima and the true optima, and to what extent the performance loss can be mitigated. Although prior work developed methods to improve the robustness of surrogate models and their associated optimization processes, a provably quantifiable relationship between an imperfect surrogate and the corresponding performance gap, as well as whether prior methods directly address it, remain elusive. To shed light on this important question, we present a theoretical framework to understand offline black-box optimization, by explicitly bounding the optimization quality based on how well the surrogate matches the latent gradient field that underlines the offline data. Inspired by our theoretical analysis, we propose a principled black-box gradient matching algorithm to create effective surrogate models for offline optimization, improving over prior approaches on various real-world benchmarks.\vspace{-3mm}
\end{abstract}

\begin{figure*}[t]
    \centering
    \includegraphics[width=\textwidth]{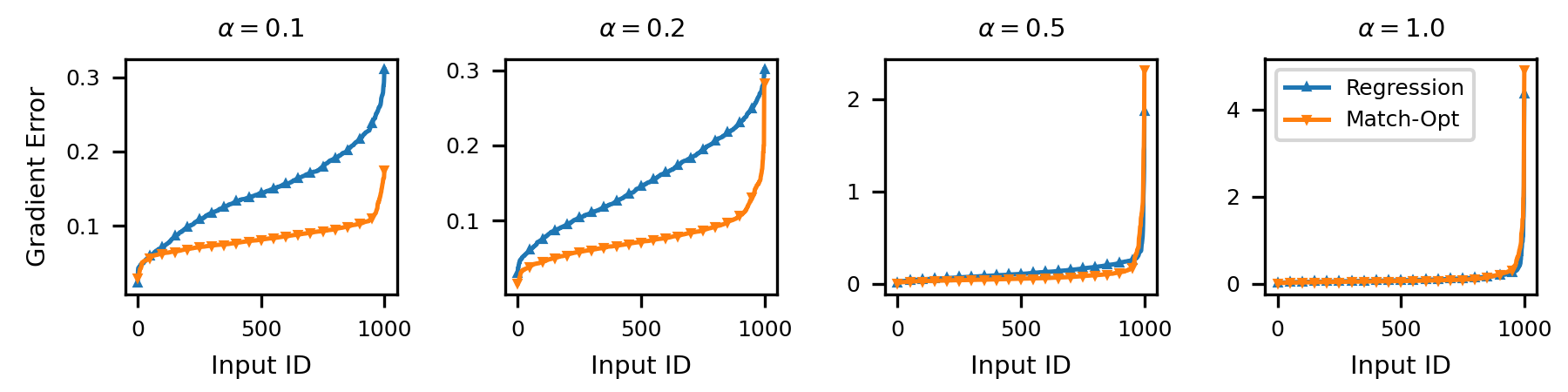}\vspace{-5mm}
    \caption{Comparison of gradient estimation error incurred by \ourmethod~(orange) and standard regression (blue) while learning the gradient field of the Shekel function on $4$-dimensional input space at different out-of-distribution (OOD) settings where test inputs were drawn from $\mathbb{N}(0, \alpha\mathbf{I})$ while training inputs were drawn from $\mathbb{N}(0, \mathbf{I})$. Smaller $\alpha$ indicates larger deviation from the offline data regime, which widens the performance gap between \ourmethod~and standard regression.}\vspace{-4mm}
    \label{fig:synthetic}
\end{figure*}

\vspace{-2.0ex}

\section{Introduction}
\label{sec:intro}
Many science and engineering applications involve optimizing an {\em expensive-to-evaluate black-box objective function} over large design spaces. Some examples include design optimization over candidate molecules, proteins~\citep{nguyen2005evolutionary}, drugs, biological sequences, and superconducting materials~\citep{si2016high}. To evaluate candidate designs, we need to perform physical lab experiments or computational simulations which are labor-intensive and impractical to do in an online manner. {\em Offline optimization} \citep{TrabuccoArXiv22,TrabuccoICML21}~is a more practical setting where we assume the access to a dataset of input and objective function evaluation pairs, and the overall goal is to use this offline training data to uncover optimal designs. 

The prototypical approach \citep{hutter2011sequential,BrookeICML19} to solve offline optimization problems is to learn a surrogate model from the given training data which can predict the objective function value for unknown inputs and find optimal input (i.e., maximizer) for this surrogate using gradient-based methods. The key implicit assumption behind this approach is that we can learn an accurate surrogate model over the entire input space using supervised learning. However, this is rarely achievable in practice due to the size and sparsity of the offline training data. In most cases, the surrogate model is only reliable within a constrained neighborhood of the offline data~\citep{ClaraNeurIPs20} and can be highly erroneous outside this neighborhood. Consequently, there will be a discrepancy between the gradient fields of the target objective function and the surrogate model which will misguide the gradient search towards sub-optimal solutions. 

This raises two related questions. {\em First}, how does the discrepancy in gradient estimation affect the performance gap between the optima of the surrogate model and the target objective function? {\em Second}, how do we learn surrogate models that closely approximate the gradient field of the target function? Both questions are challenging, given that the target function's gradient field is non-observable even at the offline training data points, and have not been studied by prior work. In fact, we note that while there is an existing literature on random gradient estimation methods~\citep{fu2015stochastic,WangAISTATS18}, those methods require the ability to actively sample data from the black-box target function which is not possible in the context of offline optimization. 

{\bf Contributions.} The main contributions of this paper include (1)~theoretically-sound answers to the above two questions; and (2)~practical demonstration of their significance on real-world offline design optimization problems:

\begin{figure*}[t]
\centering   
\includegraphics[width=0.82\textwidth]{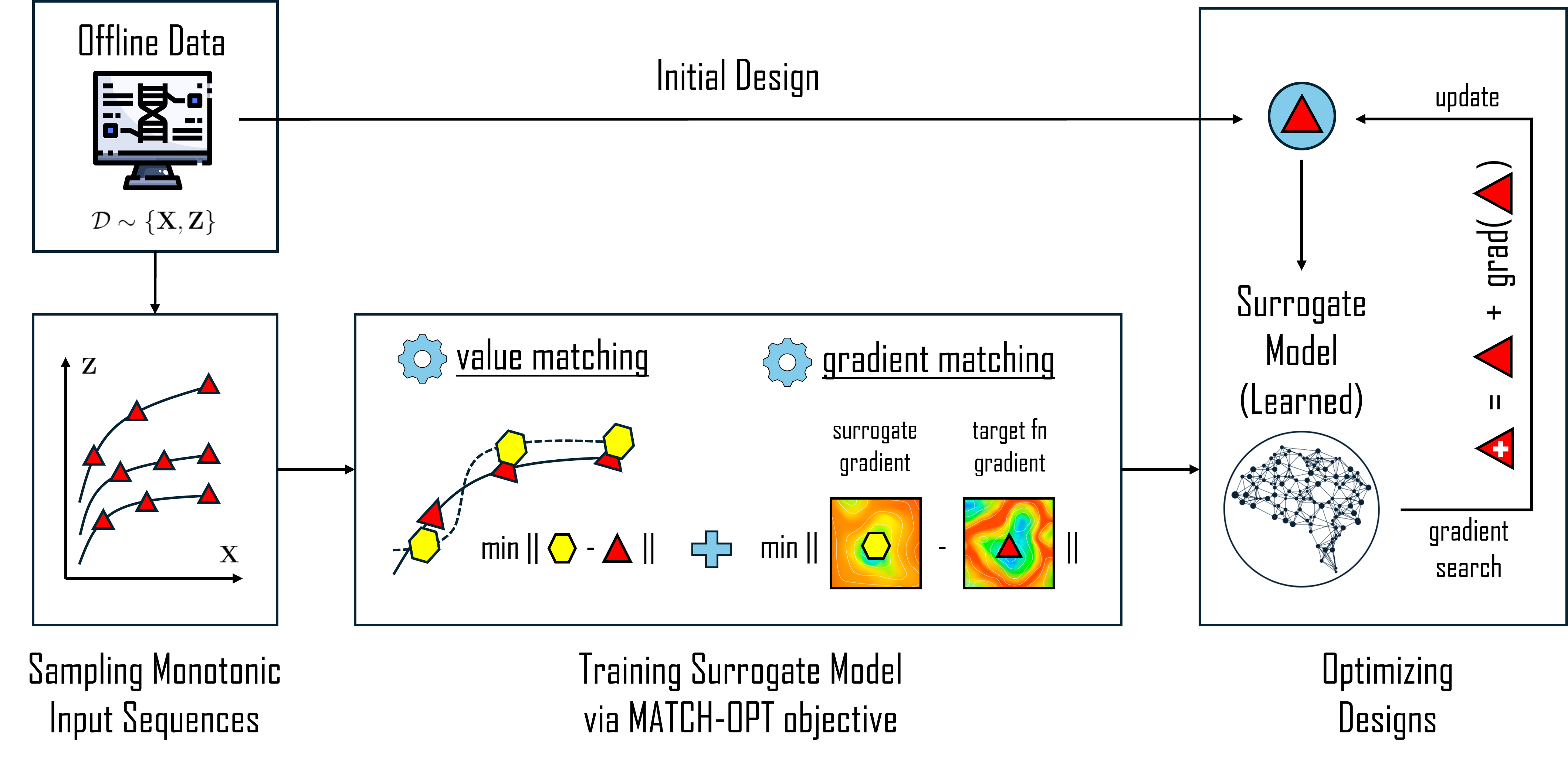}\vspace{-3mm}
\caption{Our approach \ourmethod~synthesizes input sequences with monotonically increasing target function values from the offline dataset, which are used to train a parametric surrogate model. Our loss function incorporates both standard regression loss (i.e., value matching) and a novel gradient matching loss. We perform gradient search on the trained surrogate to find optimized designs.}\vspace{-3mm}
\label{fig:workflow}
\end{figure*}

{\bf 1.} To answer the first question, we present a theoretical framework that characterizes the offline optimization performance of gradient-based search guided by a surrogate model. We provably bound the performance gap between the optima of the target function and trained surrogate as a function of how well the surrogate matches the (latent) gradient field of the target function on the offline training data. Our derived bound is non-trivial and yet model-agnostic, making it broadly applicable (Section~\ref{sec:theory}).

{\bf 2.} To answer the second question, we present a principled gradient matching algorithm, referred to as \ourmethod, that is inspired by our theoretical analysis. Intuitively, our analysis shows that the worst-case performance of an optimizer following the surrogate gradient is bounded with the gradient gap between the surrogate and target function, and that the bound is tight up to a constant with a sufficiently small learning rate. Hence, a surrogate model trained to directly match gradients will result in good offline optimization performance with gradient search from diverse starting points (referred to as ``reliable''). An overview of our algorithm is given in Fig.~\ref{fig:workflow}. Our algorithm \ourmethod~is model-agnostic and allows us to approximate the gradient field that underlies the offline training data using a parametric surrogate (Section~\ref{sec:algorithm}). In practice, existing offline optimizers exhibit high variance in their performance across diverse optimization tasks. \ourmethod~ is aimed at achieving reliable performance to address this challenge.

To provide an intuition and sanity check to readers, we visualize the reliability of our method's gradient estimation in several out-of-distribution (OOD) settings. We train our method, \ourmethod, and a standard regression model on the same set of random inputs drawn from $\mathbb{N}(0, \mathbf{I})$ and their Shekel function~\citep{molga2005test} evaluations. Fig.~\ref{fig:synthetic} plots the (sorted) gradient estimation error (i.e., the norm difference between predicted and oracle gradients) achieved by the two approaches at $1000$ random inputs drawn from different OOD distributions $\mathbb{N}(0, \alpha\mathbf{I})$ parameterized with different values of $\alpha \in [0.1, 0.2, 0.5, 1.0]$. It is observed that (1) when the test and train distributions are the same ($\alpha = 1.0$), the performance of the two approaches are the same; but (2) when $\alpha$ decreases (i.e., larger deviation from the offline data regime), our approach achieves significantly smaller error, suggesting that a direct gradient matching is more reliable in OOD data regimes. While this behaviour does not necessarily translate into better predictive accuracy, our Theorem~\ref{thm:1} demonstrates that it will indeed minimize the optimization risk as we follow the surrogate's gradient towards the goal of finding the maximum of the target objective function. We note that similar ideas have shown great empirical success in structured prediction where models were learned to guide greedy search in combinatorial spaces~\citep{JMLR:v15:doppa14a}. 

{\bf 3.} Finally, we demonstrate the efficacy of \ourmethod~on diverse real-world optimization problems from the design-bench benchmark \citep{TrabuccoArXiv22}. Our results show that \ourmethod~consistently shows improved optimization performance over existing baselines, and produces high-quality solutions with gradient search from diverse starting points (Section~\ref{sec:experiment}). Our code is publicly available at \url{https://github.com/azzafadhel/MatchOpt}.\vspace{-2mm}

\section{Background and Problem Setup}

\noindent {\bf Offline 
Black-box Optimization.} Suppose $\mathfrak{X}$ is an input space where each $\mathbf{x} \in \mathfrak{X}$ is a candidate input. Let $g: \mathfrak{X} \mapsto \Re$ be an unknown, expensive real-valued objective function which can evaluate any given input $\mathbf{x} \in \mathfrak{X}$ to produce output $z$ = $g(\mathbf{x})$. For example, in material design application, $g(\mathbf{x})$ corresponds to running a physical lab experiment. Our overall goal is to find an optimal input or design $\mathbf{x}_\ast \in \mathfrak{X}$ that maximizes the output of an experimental process $g(\mathbf{x})$,\vspace{-2mm}
\begin{eqnarray}
\mathbf{x}_\ast &\triangleq& \underset{\mathbf{x} \in \mathfrak{X}}{\argmax} \ \ 
 g(\mathbf{x}) \ . \label{eq:1}
\end{eqnarray}
We are provided with a 
dataset of $n$ input-output pairs $\mathfrak{D}$=$\{(\mathbf{x}_1, z_1), (\mathbf{x}_2, z_2), \cdots,(\mathbf{x}_n, z_n)\}$ collected offline, where $z_i = g(\mathbf{x}_i)$. We do not have access to the target objective function $g$ values on inputs outside the dataset $\mathfrak{D}$.

\noindent {\bf Surrogate Model.} We do not have access to the black-box target function $g(\mathbf{x})$ beyond the offline dataset $\mathfrak{D}$ of $n$ training examples. This allows us to learn a surrogate $g_\phi(\mathbf{x})$ for $g(\mathbf{x})$ via supervised learning.
\begin{eqnarray}
\phi &\triangleq& \underset{\phi'}{\argmin} \ \sum_{i=1}^n \ell\Bigg(g_{\phi'}\Big(\mathbf{x}_i\Big),\   z_i\Bigg) \ ,\label{eq:2}
\end{eqnarray}
where $\phi$ denotes the parameters of surrogate model and $\ell(z, z')$ denotes the loss of predicting $z$ when the true objective value is $z'$ for a given input $\mathbf{x}$. For example, $\ell(z, z') = (z - z')^2$ and $g_\phi(\mathbf{x}) = \phi^\top\mathbf{x}$. 

\noindent {\bf Gradient-based Search Procedure.} Once learned, $\phi$ is fixed and we can use $g_\phi(\mathbf{x})$ as a surrogate to approximate the optimal design as:
\begin{eqnarray}
\hspace{-5mm}\mathbf{x}_\phi &\simeq& \mathbf{x}_\phi^{m} \ \ \ \text{where}\ \ \ \mathbf{x}_\phi^{k + 1} \ \triangleq\ \mathbf{x}_\phi^{k} \ \ +\ \ \lambda \cdot \nabla g_\phi\left( \mathbf{x}_\phi^{k}\right) 
 \label{eq:3a}  
\end{eqnarray}
which is defined recursively for $0 \leq k \leq m - 1$ via a $m$-step gradient ascent process starting from an initial solution $\mathbf{x}_\phi^0 = \mathbf{x}^0$ with a fixed learning rate $\lambda > 0$. The final iterate $\mathbf{x}_\phi^m$ is referred to as the solution of gradient search guided by the surrogate. The use of a differentiable surrogate to find the optimal design also imposes the following implicit assumption on the unknown black-box target function.
\begin{assumption}
\label{assumption:1}
The target function is either differentiable or sufficiently close to a differentiable (black-box) proxy.\vspace{-2mm}
\end{assumption}
This is not an unreasonable assumption because otherwise, offline optimization and more broadly, machine learning is an ill-posed problem following a simple thought experiment:

{\em Suppose the target function is not sufficiently close to any differentiable proxy functions, no algorithm that uses a differentiable surrogate can learn a good approximation of the target function.}

To see this, suppose there exists an algorithm that uses a differentiable surrogate that can approximate well the target function. This means there exists a differentiable function that is sufficiently close to the target function, which contradicts the premise of the experiment. Thus, we argue that the offline optimization task is only meaningful within the set of target functions that can be characterized sufficiently accurately using a (black-box) differentiable proxy.  

As such, we would want the above surrogate-guided solution $x_\phi^m$ to match the solution of gradient search guided by the target function's derivative, or the derivative of a differentiable proxy function that is closest to the target function $g$. 
Henceforth, we will refer to this as the target function's gradient, and the solution guided by the gradient of the target function is defined as:
\begin{eqnarray}
\hspace{-7.5mm}\mathbf{x}_\ast &\simeq& \mathbf{x}_\ast^{m} \ \ \ \text{where}\ \ \ \mathbf{x}_\ast^{k + 1} \ \triangleq\  \mathbf{x}_\ast^{k} \ \ +\ \ \lambda \cdot \nabla g\left( \mathbf{x}_\ast^{k}\right) \label{eq:3b}
\end{eqnarray}
which forms a similar gradient search of $m$ steps with the same initial solution $\mathbf{x}_\ast^0 = \mathbf{x}_\phi^0 = \mathbf{x}^0$ and learning rate $\lambda > 0$. As such, a discrepancy between target function gradients and surrogate gradients can result in a performance gap between the objective function values of $\mathbf{x}_\ast^m$ and $\mathbf{x}_\phi^{m}$. This paper therefore studies two related questions in the context of surrogate-guided gradient search for offline optimization.

{\bf Q1.} How does the discrepancy between target function and surrogate gradients impact the quality of uncovered solutions? This will be discussed in Section~\ref{sec:theory}.

{\bf Q2.} How to learn surrogate models that can closely approximate target function gradients using the offline training data $\mathfrak{D}$? This will be discussed in Section~\ref{sec:algorithm}.

\vspace{-1.0ex}

\section{Theoretical Analysis}
\label{sec:theory}
This section provides a rigorous theoretical analysis to answer {\bf Q1}. Explicitly, we derive an upper-bound for the performance gap between gradient search guided by the target function and the trained surrogate, which is characterized explicitly in terms of how well the surrogate's gradient field fits with the offline data. 

\noindent {\bf Performance Gap.} First, we define the performance of the solution found via $m$ steps of gradient ascent on $g_\phi(\mathbf{x})$ starting from $\mathbf{x}_\phi^0 = \mathbf{x}^0$ via
\begin{eqnarray}\mathfrak{R}^m_{g_\phi}\left(\mathbf{x}^0_\phi\right) \ \ =\ \ \mathfrak{R}^m_{g_\phi}\left(\mathbf{x}^0\right) &=& g(\mathbf{x}_\ast) \ -\ g(\mathbf{x}_\phi) \nonumber\\
&=& g(\mathbf{x}_\ast) \ -\ g(\mathbf{x}_\phi^m) \ .\label{eq:4}    
\end{eqnarray}
where $\mathbf{x}_\phi^m$ is defined in \eqref{eq:3a}. Similarly, we have $\mathfrak{R}^m_g(\mathbf{x}^0_\ast) = \mathfrak{R}^m_g(\mathbf{x}^0) = g(\mathbf{x}_\ast) - g(\mathbf{x}_\ast^m) \geq 0$. Note that we are distinguishing between the solution $\mathbf{x}_\ast^m$ and $\mathbf{x}_\ast$ here because, often, finding $\mathbf{x}_\ast$ is intractable even with access to the target function $g(\mathbf{x})$ (e.g., combinatorial spaces). Thus, it is more practical to compare the surrogate solution with the solution found via following the target function's gradient, rather than the true optima. We can now define the performance gap and state our main result.

\begin{definition}
\label{def:1}
For a fixed gradient ascent process starting from $\mathbf{x}$ with $m$ update steps and learning rate $\lambda > 0$, the performance gap between the surrogate solution $\mathbf{x}_\phi^m$ and the target function solution $\mathbf{x}_\ast^m$, $\mathfrak{G}_{m,\lambda}(\mathbf{x})$, is given by: 
\begin{eqnarray}
\mathfrak{G}_{m,\lambda}(\mathbf{x}) &\triangleq& \Big\|\ \mathfrak{R}^m_g(\mathbf{x}) \ -\  \mathfrak{R}^m_{g_\phi}(\mathbf{x})\ \Big\| \ ,
\label{eq:5}
\end{eqnarray}
where $\mathfrak{R}_g$ and $\mathfrak{R}_{g_\phi}$ are as defined above.
\end{definition}

\begin{theorem}[Worst-case optimization risk bound in terms of gradient estimation error]
\label{thm:1}
Suppose $g(\mathbf{x})$ is a $\ell$-Lipschitz continuous and $\mu$-Lipschitz smooth function. The worst-case performance gap, $\mathfrak{G}_{m, \lambda} \triangleq \max_{\mathbf{x}}\mathfrak{G}_{m,\lambda}(\mathbf{x})$, between $g$ and some arbitrary surrogate $g_{\phi}$ is upper-bounded by:
\begin{eqnarray}
\mathfrak{G}_{m, \lambda}&\leq& m\lambda\ell\Big(1 + \lambda\mu\Big)^{m - 1} \nonumber\\
\hspace{-2mm}&\times&\hspace{-2mm} \max_{\mathbf{x}}\Big\|\nabla g(\mathbf{x}) - \nabla g_\phi(\mathbf{x})\Big\| \ . \label{eq:7}
\end{eqnarray}
Note that despite the exponential dependence on $m$, the bound becomes tight and independent of $m$ when the learning rate $\lambda \leq 1/m$, which is the case in all our experiments. See Appendix A for a detailed derivation.
\end{theorem}

\noindent Theorem~\ref{thm:1} establishes that the worst-case performance gap between the surrogate and target function solutions is upper-bounded by the maximum norm difference between the surrogate and target function gradients over the input space. This provides a direct quantification of optimization quality as a function of gradient discrepancies. In addition, the result of Theorem~\ref{thm:1} also characterizes a balance between the risk and potential of gradient search in terms of the learning rate and the number of update steps. 

\noindent As the learning rate $\lambda$ or the number of search steps $m$ approaches zero, the bound in Theorem~\ref{thm:1} also approaches zero. This means an extremely conservative gradient search (one that barely moves) would minimize the gap between $\mathfrak{R}_{g_\phi}$ and $\mathfrak{R}_g$, making the performance of the surrogate solution arbitrarily close to that of the target function solution. However, such a conservative strategy would also widen the gap between the target function solution and the true optima, and thus will ultimately deteriorate the overall performance of offline optimization. Conversely, an explorative search that uses larger $\lambda$ and $m$ would bring $\mathfrak{R}_g$ closer to zero, making the target function solution arbitrarily close to the true optima. Simultaneously, it also widens the gap between the surrogate and target function solution, again reducing the performance of offline optimization. Furthermore, as the bound in Eq.~\eqref{eq:7} holds for all possible choices of $g_\phi$, we can tighten it with respect to $g_\phi$. That is:\vspace{-1mm}
\begin{eqnarray}
\mathfrak{G}_{m,\lambda} &\leq& m\lambda\ell\Big(1 + \lambda\mu\Big)^{m - 1} \nonumber\\
&\times& \min_{\phi}\max_{\mathbf{x}}\Big\|\nabla g(\mathbf{x}) - \nabla g_\phi(\mathbf{x})\Big\| \ . \label{eq:8} 
\end{eqnarray}
For a fixed gradient based search configuration $(m, \lambda)$, the offline optimization task is therefore reduced to solving a minimax program,\vspace{-1mm}
\begin{eqnarray}
\phi_\ast &=& \argmin_{\phi} \max_{\mathbf{x}} \Big\|\nabla g(\mathbf{x}) - \nabla g_\phi(\mathbf{x})\Big\| \ ,\label{eq:9}
\end{eqnarray}
which is non-trivial since we do not have direct access to $\nabla g(\mathbf{x})$. Instead, we only have the value of $g(\mathbf{x}_i)$ at a finite number of inputs $\{\mathbf{x}_i\}_{i=1}^n$. Fortunately, this can be circumvented via matching the gradient $\nabla g_\phi(\mathbf{x})$ to that of the observational data, which approximately represents the target function. This is detailed in Section~\ref{sec:algorithm} below.

\noindent {\bf Remark.} As mentioned in the statement of the theorem, the exponential dependence on $m$ of the above bound can be mitigated by choosing $\lambda \leq 1/m$. To see this, note that
\begin{eqnarray}
\hspace{-6mm}\left(1 + \lambda \cdot \mu\right)^{m-1} &\leq& \left(1 + \frac{\mu}{m}\right)^{m-1} \ \ <\ \ \left(1 + \frac{\mu}{m}\right)^m
\end{eqnarray}
which will approach $e^\mu$ in the limit of $m$. Here, we use the known fact that $\mathrm{lim}_{m\rightarrow\infty}(1 + \mu/m)^m = e^\mu$ with $\mu > 0$. As such, when $m$ is sufficiently large, the bound in Theorem~\ref{thm:1} is upper-bounded with $m \cdot\lambda \cdot \ell \cdot (1 + \lambda \cdot \mu)^{m - 1} \cdot \text{\em gradient-gap} \simeq \ell \cdot e^\mu \cdot\text{\em gradient-gap}$ which asserts that the worst-case performance gap of our offline optimizer is approaching (in the limit of $m$) $\ell \cdot e^\mu \cdot \max_{\mathbf{x}}\|\nabla g(\mathbf{x}) - \nabla g_\phi(\mathbf{x})\| = \mathbf{O}(\max_{\mathbf{x}}\|\nabla g(\mathbf{x}) - \nabla g_\phi(\mathbf{x})\|)$ which is not dependent on the no. of gradient steps.

\begin{algorithm}[ht]
    \caption{\ourmethod: Black-Box Gradient Matching from Offline Training Data}
    \footnotesize
   \textbf{Input}: Dataset $\mathfrak{D}=\{(\mathbf{x}_i,z_i)\}_{i=1}^{n}$, initial surrogate model parameters $\phi$, length of monotonic synthetic paths $m$, number of iterations $\tau$, learning rate $\lambda > 0$\\
    \textbf{Output}: Surrogate $g_{\phi}$ with parameters $\phi^{(\tau)}$
    \begin{algorithmic}[1]
        \STATE Generate monotonic trajectories $\mathcal{C}^m$ via strategy from~\citet{krishnamoorthy2023generative},~\citet{kumar19}
        \STATE $\phi^{(0)} \leftarrow \phi$ // initialize parameters of surrogate model
        \FOR{\(t \gets 0: \tau-1\)}
        \STATE $\mathfrak{L} \leftarrow 0$ // initialize the average loss
        \FOR{$\zeta = (\mathbf{x}_1, \ldots, \mathbf{x}_m) \in \mathcal{C}^m$}
        \STATE $\displaystyle\mathfrak{L}^\zeta_g \leftarrow$ gradient matching loss using Eq.~\ref{eq:13} with $\phi = \phi^{(t)}$
        \STATE $\displaystyle\mathfrak{L}^{\zeta}_r \leftarrow \alpha \cdot \sum_{i=1}^m\Big(g\left(\mathbf{x}_i\right) - g_{\phi}\left(\mathbf{x}_i\right)\Big)^2$ // using $\phi = \phi^{(t)}$
        \STATE $\displaystyle\mathfrak{L} \leftarrow \mathfrak{L} + \Big|\mathcal{C}^m\Big|^{-1}\Big(\mathfrak{L}^\zeta_g + \mathfrak{L}^\zeta_r\Big)$ // update average loss
        \ENDFOR
        \STATE $\phi^{(t+1)} \leftarrow \phi^{(t)} + \lambda \cdot \nabla_\phi \mathfrak{L}\Big|_{\phi = \phi^{(t)}}$ 
        \ENDFOR
        \STATE \textbf{return} the learned surrogate model $g_{\phi}$ with $\phi = \phi^{(\tau)}$
    \end{algorithmic}
    {\label{alg:training}}
\end{algorithm}

\section{Practical Algorithm: \ourmethod}
\label{sec:algorithm}

This section answers {\bf Q2} by providing a principled algorithm for black-box gradient matching, which we name \ourmethod. The crux of solving Eq.~\eqref{eq:9} lies with how we approximate the target function's gradient field when we are given evaluations of the target function at a fixed set of inputs (i.e., offline dataset). Previous approaches often address this by sampling perturbed values around a chosen input and use the finite difference method to approximate its gradient~\citep{fu2015stochastic,WangAISTATS18}. However, these methods require the ability to query the target function for evaluations of perturbed data points, which is not possible in the offline optimization setting. To overcome this challenge, we  leverage the fundamental theorem for line integrals, which states that for any two inputs $\mathbf{x}$ and $\mathbf{x}'$ with corresponding values $z = g(\mathbf{x})$ and $z' = g(\mathbf{x}')$:
\begin{eqnarray}
\hspace{-6mm}\Delta z \ \triangleq \ z - z' \hspace{-2mm}&=&\hspace{-2mm} \Big(\mathbf{x}' - \mathbf{x}\Big)^\top \int_0^1\Big[\nabla g\Big(h(t) \Big)\Big]\mathrm{d}t\nonumber \\
\hspace{-2mm}&\simeq&\hspace{-2mm}
\Big(\mathbf{x}' - \mathbf{x}\Big)^\top \int_0^1\Big[\nabla g_\phi\Big(h(t) \Big)\Big]\mathrm{d}t \ ,
\label{eq:10}  
\end{eqnarray}
where $h(t) = \mathbf{x} \cdot (1 - t) + \mathbf{x}' \cdot t$. The above approximation holds when $\nabla g_\phi$ closely estimates $\nabla g$. To enforce this, we therefore need to find $\phi$ such that the averaged difference between the LHS and RHS of \eqref{eq:10} is minimized. This is achieved by solving $\phi^\ast = \argmin_\phi \mathfrak{L}_g(\phi)$, where
\begin{eqnarray}
\hspace{-6mm}\mathfrak{L}_g(\phi) \hspace{-2mm}&\triangleq&\hspace{-2mm}\mathbb{E}_{\mathbf{x},\mathbf{x}'\in\mathfrak{D}}\Bigg(\Delta z - \Delta\mathbf{x}^\top\hspace{-1mm}\int_0^1 \nabla g_\phi\Big(h(t)\Big)\mathrm{d}t\Bigg)^2 .\label{eq:11}
\end{eqnarray}
Eq.~\eqref{eq:11} provides a tractable learning objective via taking the empirical expectation over random inputs sampled from the offline training dataset $\mathfrak{D}$. It can also be shown that minimizing Eq.~\eqref{eq:11} will indeed decrease the gradient gap between the surrogate and the black-box target function (see Appendix~\ref{app:gradient-gap} for a detailed derivation). Thus, by virtue of Theorem~\ref{thm:1}, minimizing Eq.~\eqref{eq:11} has the effect of decreasing the upper-bound on the worst-case performance of offline optimization using the learned surrogate's gradient. 


Furthermore, we note that in the ideal scenario, Eq.~\eqref{eq:11}~can be indirectly solved using a regression approach (i.e., value matching) because the gradient fields of $g$ and $g_\phi$ must be the same when $g_\phi(\mathbf{x})$ accurately estimates $g(\mathbf{x})$ for every input $\mathbf{x}$. However, as long as there are discrepancies, it is unclear which surrogate gradient (among surrogate candidates that approximate the target function equally well) would minimize the gradient discrepancy. As such, we argue that a direct gradient matching approach is more preferable in this case. This statement is supported by both our synthetic experiment (see Fig.~\ref{fig:synthetic}) and real-world experiments presented in Section~\ref{sec:results}. 

{\bf Practical Considerations.} A na\"ive optimization of Eq.~\eqref{eq:11} requires enumerating over all pairs of training inputs, which is more expensive than a standard regression algorithm. To avoid this overhead, we adopt the strategy of~\citet{krishnamoorthy2023generative} and~\citet{kumar2020model}, which organizes training data into trajectories of monotonically increasing target function values. These trajectories mimic realistic optimization paths and thus encourages the model to learn the behavior of a gradient-based optimization algorithm, and thus allows the gradient matching algorithm to focus more on strategic input pairs that are more relevant for gradient estimation.

\noindent Specifically, let $\mathcal{C}^m$ denote a finite set of $m$-hop synthetic input paths with increasing objective function values, such that if $\zeta = \{\mathbf{x}_1, \mathbf{x}_2, \ldots, \mathbf{x}_m\} \in \mathcal{C}^m$, we have $g(\mathbf{x}_{i+1}) \geq g(\mathbf{x}_i)$. To sample trajectories from this set, we first bin the offline inputs based on their percentiles in the dataset, and subsequently sample one input from each bin to form a trajectory with monotonically increasing function values. We adapt the loss function in Eq.~\eqref{eq:11} to optimize along the sampled paths, and thus focus on estimating gradient information that is relevant to the downstream search procedure. That is, we aim to minimize $\mathfrak{L}_{g}(\phi; \mathcal{C}^m) \triangleq \mathbb{E}_{\zeta\in\mathcal{C}^m}[\mathfrak{L}_{g}(\phi; \zeta)]$, where:
\begin{eqnarray}
\hspace{-5mm}\mathfrak{L}_{g}\left(\phi; \zeta\right)\hspace{-2mm}&\triangleq&\hspace{-2mm}\sum_{i=1}^{m-1}\Bigg(\Delta z - \Delta\mathbf{x}^\top \int_0^1 \nabla g_\phi\Big(h_i(t)\Big)\mathrm{d}t\Bigg)^2 \nonumber\\
\hspace{-2mm}&\simeq&\hspace{-2mm} \sum_{i=1}^{m-1}\Bigg(\Delta z -\frac{1}{2\kappa} \sum_{u=1}^{\kappa} \Bigg(\Delta\mathbf{x}^\top\Big(\mathbf{r}_i(u)\Big)\Bigg)\Bigg)^2 
\label{eq:13}
\end{eqnarray}
with $\mathbf{r}_i(u) = \nabla g_\phi(h_i((u - 1) / \kappa)) + \nabla g_\phi(h_i(u / \kappa))$ and $h_i(t) = \mathbf{x}_i \cdot (1 - t) + \mathbf{x}_{i+1} \cdot t $. Here, Eq.~\eqref{eq:13} takes empirical expectation over the successive pairs along the synthesized trajectories $\zeta \in \mathcal{C}^m$, $\Delta z \triangleq g(\mathbf{x}_{i+1}) - g(\mathbf{x}_i)$, and $\Delta\mathbf{x} \triangleq \mathbf{x}_{i+1} - \mathbf{x}_i$. In addition, the integral inside the expectation on the RHS of Eq.~\eqref{eq:13} is approximated via a discretization of $(0,1)$ into $\kappa$ intervals with equal lengths. Our empirical investigations suggest that setting $\kappa=5$ works best in practice.
Finally, our ultimate loss function $\mathfrak{L}(\phi)$ combines Eq.~\eqref{eq:13} with the regression loss along the synthetic trajectory to achieve the best of both worlds:
\begin{eqnarray}
\hspace{-6mm}\mathfrak{L}(\phi) \hspace{-2mm}&\triangleq&\hspace{-2mm}   \mathfrak{L}_{g,\mathcal{C}^m}(\phi) +  \mathbb{E}_{\zeta\in\mathcal{C}^m}\sum_{i=1}^{m} \Bigg(g(\mathbf{x}_i) - g_\phi(\mathbf{x}_i)\Bigg)^2\label{eq:14}
\end{eqnarray}
See Algorithm~\ref{alg:training} for a complete pseudo-code of our method. Interestingly, Eq.~\eqref{eq:14} is also motivated by a modification of the above theoretical analysis in Section~\ref{sec:theory}, which characterizes a condition under which the worst-case optimization risk in Theorem~\ref{thm:1} has a tighter bound. It can be shown that such a bound will depend on both the (worst-case) gradient- and value-matching quantities, which inspires the addition of the above regression loss in Eq.~\eqref{eq:14} to the original loss in Eq.~\eqref{eq:13}. This is formalized in Theorem~\ref{thm:1b} below.

\begin{theorem}[Generalized worst-case optimization risk bound]
\label{thm:1b}
Suppose the target objective function $g(\mathbf{x})$ is a $\ell$-Lipschitz continuous and $\mu$-Lipschitz smooth function. For all $a \in (0, 1)$, the worst-case performance gap, $\mathfrak{G}_{m, \lambda} \triangleq \max_{\mathbf{x}}\mathfrak{G}_{m,\lambda}(\mathbf{x})$, between $g$ and some arbitrary surrogate $g_{\phi}$ with Lipschitz constant $\ell_\phi$ is upper-bounded by:
\begin{eqnarray}
\mathfrak{G}_{m, \lambda} &\leq& m \cdot 2a \cdot \max_{\mathbf{x}} \Big\|g(\mathbf{x}) - g_\phi(\mathbf{x})\Big\| \nonumber\\
&+& m \cdot \Big(\ell + a \cdot (\ell_\phi - \ell)\Big) \cdot \Big(1 + \lambda \mu\Big)^{m-1} \nonumber\\
&\times& \max_{\mathbf{x}} \Big\|\nabla_{\mathbf{x}}g(\mathbf{x}) - \nabla_{\mathbf{x}}g_\phi(\mathbf{x})\Big\| \ ,
\end{eqnarray}
which is tighter than the bound in Theorem~\ref{thm:1} when the Lipschitz constant $\ell_\phi$ of the surrogate satisfies:
\begin{eqnarray}
\ell_\phi \hspace{-2mm}&\leq& \hspace{-2mm} \lambda \ell -  \frac{2 \cdot \max \Big\|g(\mathbf{x}) -  g_\phi(\mathbf{x})\Big\|}{\Big(1 + \lambda \mu\Big)^{m-1} \cdot \max \Big\|\nabla_{\mathbf{x}}g\big(\mathbf{x}\big) - \nabla_{\mathbf{x}}g_\phi\big(\mathbf{x}\big)\Big\|} \nonumber
\end{eqnarray}
The detailed proof of this result is deferred to Appendix~\ref{app:regularizer-analysis}.
\end{theorem}
Although it has not been investigated how to further condition the training loss in Eq.~\eqref{eq:14} so that $\ell_\phi$ satisfies the above, we are able to empirically demonstrate the benefit of adding the regression loss along the synthetic monotonic trajectories via an ablation study in Section~\ref{sec:results}. 

{\bf Complexity Analysis.} Given a $m$-hop synthetic sequence $\zeta$ of $d$-dimensional inputs, each step of the inner loop in Algorithm~\ref{alg:training} will require a linear scan over $m$ segments. For each segment, the algorithm needs to compute (1) the gradient matching loss, which costs $\mathcal{O}(dm\kappa |\phi|)$ where $\kappa$ is the granularity of the discretization in~\eqref{eq:13} and $|\phi|$ is the number of parameters of the surrogate model, and (2)  the regression regularizer on this path, which costs $\mathcal{O}(m|\phi|)$. Thus, suppose $p = |\mathcal{C}^m|$ synthetic input sequences/paths were generated for our algorithm, the entire inner loop of Algorithm~\ref{alg:training} will incur a total cost of $\mathcal{O}(p \cdot (dm\kappa |\phi| + m|\phi|)) = \mathcal{O}(p \cdot dm\kappa |\phi|)$. This is the complexity per training iteration. For $\tau$ iterations, the total complexity of Algorithm~\ref{alg:training} will be $\mathcal{O}(\tau \cdot p \cdot dm\kappa |\phi|)$.

\vspace{-1.0ex}

\section{Experiments}
\label{sec:experiment}
This section describes the set of benchmark tasks used to evaluate and compare the performance of \ourmethod~with those of other baselines (Section~\ref{sec:benchmark}), the configurations of both our proposed algorithm and those baselines (Section~\ref{sec:config}), as well as their reported results (Section~\ref{sec:results}).

\vspace{-1.0ex}

\subsection{Benchmarks} 
\label{sec:benchmark}
Our empirical studies are conducted on six benchmark tasks from a diverse set engineering domains. Each task comprises a black-box target function and an offline training dataset, which is a small subset of a much larger dataset used to train the target function. Each participating algorithm only has access to the offline dataset. The target function is only used to evaluate the performance of the final inputs recommended by those offline optimizers. The specifics of these datasets and their target functions are further provided in the design baseline package~\citep{TrabuccoArXiv22}. Four tasks are defined over continuous input spaces, whereas the other two are discrete, as summarized below.

{\bf 1 \& 2.} The Ant Morphology~\citep{brockman2016openai} (\textsc{Ant}) and D'Kitty Morphology dataset~\citep{ahn2020robel} (\textsc{DKitty}) collect morphological observations of two robots and their corresponding rewards in moving as fast as possible, or towards a specific location. The 
parameters of the robot is defined over a 60/56-dimensional continuous 
space. 

{\bf 3.} The Hopper Controller dataset~\citep{ahn2020robel} (\textsc{Hopper}) collects observations of a neural network policy weights and their rewards on the Hopper-v2 locomotion task in OpenAI Gym~\citep{brockman2016openai}. The search space is defined over 5126-dimensional continuous space.

{\bf 4.} The Superconductor dataset~\citep{BrookeICML19} (\textsc{SCon}) collects observations of superconductor molecules and their critical temperatures. Each molecule is represented by a 86-dimensional continuous vector. 

{\bf 5 \& 6.} The TF-Bind-8 (\textsc{Tf8}) and TF-Bind-10 (\textsc{Tf10}) datasets~\citep{tfbind8} collect the binding activity scores between a given human transcription factor and various DNA sequences of length $8$ and $10$. The goal of these \textit{discrete} tasks is to find a DNA sequence that maximizes the binding score with the given transcription factor.
\begin{table*}[ht]
\centering
\begin{tabular}{llllllll} 
\toprule
\textsc{Method} & \textsc{Ant} & \textsc{DKitty} & \textsc{Hopper} & \textsc{SCon} & \textsc{Tf8} & \textsc{Tf10} & \textsc{MNR} \\
\midrule
\textsc{GA} & 0.271 & 0.895 & 0.780 & 0.699 & 0.954 & 0.966 & 0.600\\
\textsc{Ens-Mean} & 0.517 & 0.899 & 1.524 & 0.716 & 0.926 & \textbf{0.968} & 0.500\\
\textsc{Ens-Min} & 0.536 & 0.908 & 1.420 & 0.734 & 0.959 & 0.959 & 0.467\\
\textsc{CMA-ES} & \textbf{0.974} & 0.722 & 0.620 & \textbf{0.757} & \textbf{0.978} & 0.966 & 0.367\\
\textsc{MINS} & 0.910 & 0.939 & 0.150 & 0.690 & 0.900 & 0.759 & 0.700\\
\textsc{CbAS} & 0.842 & 0.879 & 0.150 & 0.659 & 0.916 & 0.928 & 0.733\\
\textsc{RoMA} & 0.832 & 0.880 & 2.026 & 0.704 & 0.664 & 0.820& 0.667\\
\textsc{BONET} & 0.927 & 0.954 & 0.395 & 0.500 & 0.911 & 0.756& 0.683\\
\textsc{COMS} & 0.885 & 0.953 & \textbf{2.270} & 0.565 & 0.968 & 0.873 & 0.467\\
\midrule
\makecell{\textbf {\ourmethod}} & 0.931 (2) & \textbf{0.957 (1)} & 1.572 (3) & 0.732 (3) & 0.977 (2) & 0.924 (6) & \textbf{0.283 (1)}\\
\bottomrule
\end{tabular}
\caption{Performance of \ourmethod~and other baselines at $100^{\text{th}}$ percentile level. The last column shows the mean normalized rank (\textsc{MNR}) computed across all tasks (smaller is better). The individual rank of \ourmethod~on each task is included next to its performance.}
\label{tab:1}
\end{table*}

\begin{table*}[ht]
\centering
\begin{tabular}{llllllll} 
\toprule
\textsc{Method} & \textsc{Ant} & \textsc{DKitty} & \textsc{Hopper} & \textsc{SCon} & \textsc{Tf8} & \textsc{Tf10} & \textsc{MNR} \\
\midrule
\textsc{GA} & 0.130 & 0.742 & 0.089 & 0.641 & 0.510 & 0.794 & 0.600\\
\textsc{Ens-Mean} & 0.192 & 0.791 & 0.209 & 0.644 & 0.529 & \textbf{0.796} & 0.433\\
\textsc{Ens-Min} & 0.190 & 0.803 & 0.166 & \textbf{0.672} & 0.490 & 0.794 & 0.500\\
\textsc{CMA-ES} & -0.049 & 0.482 & -0.033 & 0.590 & 0.592 & 0.786 & 0.683\\
\textsc{MINS} & 0.614 & 0.889 & 0.088 & 0.414 & 0.420 & 0.465 & 0.650\\
\textsc{CbAS} & 0.376 & 0.757 & 0.013 & 0.099 & 0.442 & 0.613 & 0.817\\
\textsc{RoMA} & 0.448 & 0.760 & 0.370 & 0.420 & 0.560 & 0.780 & 0.533\\
\textsc{BONET} & \textbf{0.620} & \textbf{0.897} & 0.390 & 0.470 & 0.505 & 0.465 & 0.417\\
\textsc{COMS} & 0.557 & 0.879 & 0.379 & 0.414 & \textbf{0.652} & 0.606 & 0.467\\
\midrule
\makecell{\textbf \ourmethod} & 0.611 (3) & 0.887 (3) & \textbf{0.393 (1)} & 0.439 (6) & 0.594 (2) & 0.720 (6) & \textbf{0.350 (1)}\\
\bottomrule
\end{tabular}
\caption{Performance of \ourmethod~and other baselines at $50^{\text{th}}$ percentile level. The last column shows the mean normalized rank (\textsc{MNR}) computed across all tasks (smaller is better). The individual rank of \ourmethod~on each task is included next to its performance.}
\label{tab:2}
\vspace{-2mm}
\end{table*}

\subsection{Algorithm Configuration and Evaluation}
\label{sec:config}
{\bf Baselines.} We evaluate and compare the performance of \ourmethod~against those of multiple state-of-the-art baselines including \textsc{COMs}~\citep{TrabuccoICML21}, \textsc{RoMA}~\citep{YuNeurIPs21}, \textsc{BONET}~\citep{krishnamoorthy2023generative}. Several other baselines from the design bench benchmark~\citep{TrabuccoArXiv22} including Gradient Ascent (\textsc{GA}), Gradient Ascent Ensemble Mean (\textsc{Ens-Mean}), Gradient Ascent Ensemble Min (\textsc{Ens-Min}), covariance matrix adaptation evolution strategy (\textsc{CMA-ES})~\citep{hansen2006cma}, model inversion networks (\textsc{MINS})~\citep{kumar2020model}, conditioning by adaptive sampling (\textsc{CbAS})~\citep{BrookeICML19} are also included for a thorough comparison. The same neural network architecture is used for all baselines. Our implementation of the \ourmethod~framework is released at \url{https://github.com/azzafadhel/MatchOpt}. Other details of our experiments are deferred to Appendix~\ref{appendix:training_eval}.

{\bf Evaluation Methodology.} We follow the widely adopted evaluation methodology introduced by~\citet{TrabuccoArXiv22}. That is, each algorithm starts the search from the same initial set of $n = 128$ offline inputs and generates the corresponding set of solution candidates which are evaluated by the oracle function. For each algorithm, these ($128$) solutions are then sorted in increasing order, and the corresponding values at the $100^{\text{th}}$ percentile (maximum solution) and $50^{\text{th}}$ (median solution) are reported in Table~\ref{tab:1} and Table~\ref{tab:2} below. All target function values are normalized using the maximum and minimum values from a larger unobserved dataset (that was used to train the target function). We run each algorithm on each task four times and report the mean. 
We report their standard deviations in Appendix~\ref{app:stddev}.

{\bf Comparison Metrics.} The overall performance of a baseline against other methods across different optimization tasks can be assessed using (a)~their mean (normalized) performance; and (b)~their mean (normalized) performance rank. While the first metric has often been used in prior work, it does not account for the variation in performance among tasks. For example, normalized performance are often close to $1$ for easy tasks, whereas for harder tasks, they can be closer to $0$. The mean performance metric therefore might favor algorithms that do well on easy tasks, but poorly on other hard tasks. To mitigate such biased assessment, we consider the mean normalized rank (MNR) metric that is agnostic to such variation of performance:
\begin{eqnarray}
\mathrm{MNR}(\mathcal{A}) &\triangleq& \frac{1}{p}\sum_{i=1}^p \frac{\mathrm{rank}(\mathcal{A}; \mathrm{task}_i)}{\text{$\#$ algorithms}} \label{eq:MNR}
\end{eqnarray}
where $p$ is the number of tasks and $\mathrm{rank}(\mathcal{A}; \mathrm{task}_i) = q$ means $\mathcal{A}$ is the $q$-best algorithm for the $i$-th task. To scale the MNR to the same range of $(0,1)$ (for convenience), we also normalize the rank by the number of participating algorithms in the ranking order. An algorithm with low MNR therefore has more reliable performance across tasks, and is preferable to other methods with higher MNR. 

\vspace{-1.0ex}

\subsection{Results and Discussion}
\label{sec:results}
To demonstrate the effectiveness of \ourmethod, we report the $100^{\text{th}}$ and $50^{\text{th}}$ percentile results of all methods in Table \ref{tab:1} and Table \ref{tab:2}. 
Other than the algorithm's individual performance reported for each task, we calculate its mean normalized rank (see Eq.~\eqref{eq:MNR}) to account for the reliability of its performance (across tasks) in the comparison. 

{\bf Mean Rank Comparison.} Overall, no algorithm performs best in more than two task domains due to the diverse and challenging nature of the benchmark tasks. In fact, for the $100$-percentile performance reported in Table~\ref{tab:1}, each algorithm only performs best in at most one task. Among these, \ourmethod~performs best on the \textsc{DKitty} dataset, and second best on \textsc{Ant} and \textsc{Tf8} datasets. \ourmethod~is consistently among the top-$3$ performers on four out of six task domains, which is an evidence of its reliable performance. In fact, this is best reflected in terms of the mean normalized rank metric (\textsc{MNR}) which averages the normalized rank of each baseline across all six tasks (see Eq.~\eqref{eq:MNR}). Among all algorithms, \ourmethod~achieves the lowest MNR, which is markedly lower than that of the second lowest MNR of \textsc{COMS}. At $50^{\text{th}}$ percentile, Table~\ref{tab:2} shows that \ourmethod~achieves the best \textsc{MNR} among the baselines.

\begin{figure*}[h!]
\centering
\begin{tabular}{cc}
\includegraphics[width=0.44\textwidth]{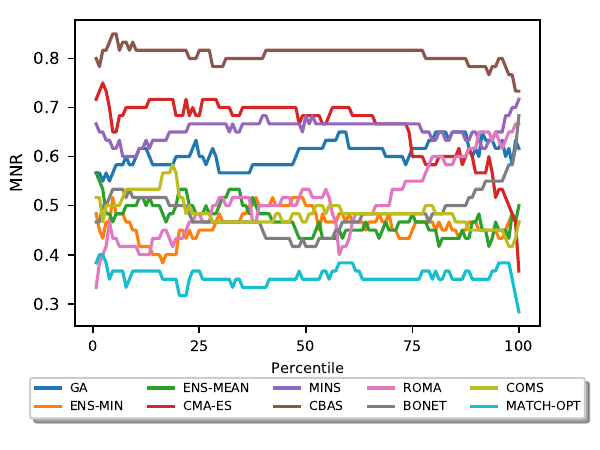} & \hspace{-3mm}\includegraphics[width=0.44\textwidth]{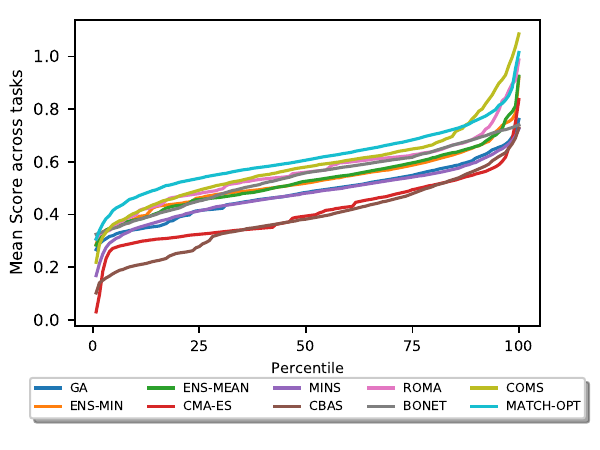} \\
(a) Mean Normalized Rank & \hspace{-3mm}(b) Mean Performance\vspace{-3mm}
\end{tabular}
\caption{Plots of (a) mean normalized ranks (MNRs); and (b) mean (normalized) performance of baselines at all performance percentiles.} 
\label{fig:mnr_mean_score}
\vspace{-3mm}
\end{figure*}

{\bf Reliability Assessment.} To further demonstrate the consistent reliability of \ourmethod~as previously alluded to in the introduction section, we also plot the \textsc{MNRs} of all competing baselines at every solution percentile level in Fig.~\ref{fig:mnr_mean_score}a. As expected, \ourmethod~achieves the lowest \textsc{MNR} at almost every percentile, averaging at approximately $0.35$ which is again markedly lower than the second lowest \textsc{MNR}. In addition, we also plot the mean performance of the tested algorithms across all percentile level in Fig.~\ref{fig:mnr_mean_score}b, which also show that \ourmethod~is the best performer (on average) between $0$- and $80$-percentile. Above that, between $80$- and $100$-percentile level, \ourmethod~ is the second best performer. The above observations (both \textsc{MNR} and mean performance) suggest that \ourmethod~is consistently the most reliable among all optimizers. We also refer the readers to Appendix~\ref{app:rankdist} which further visualizes the entire rank distribution of the tested algorithm across different percentile level. All observations are consistent with our above observations in Fig.~\ref{fig:mnr_mean_score}.

{\bf Ablation Studies for Regression Regularizer.} To demonstrate the effectiveness of our practical consideration mentioned in Section~\ref{sec:algorithm}, we conduct an ablation study comparing two versions of \ourmethod~using the original gradient matching loss in~\eqref{eq:11} (referred as \ourmethod~(no-regularizer)) and an augmented version with regression regularizer along a set of sampled synthetic input sequences in ~\eqref{eq:14} (referred to as \ourmethod~(with-regularizer)). Table~\ref{tab:3} and~\ref{tab:4} below reports the performance of these ablated methods at the $100^{\text{th}}$ and $50^{\text{th}}$ percentile of solutions respectively. Overall, we observe that \ourmethod~(with-regularizer) outperforms \ourmethod~(no-regularizer) on 4/6 tasks for both the $100^{\text{th}}$-percentile and $50^{\text{th}}$-percentile metric, thus confirming that it is an effective strategy to prioritize optimizing the gradient matching loss along critical trajectories of inputs.
\begin{table*}[ht]
\centering
\begin{tabular}{llllllr} 
\toprule
\hspace{-2mm}\textsc{Method} & \textsc{Ant} & \textsc{DKitty} & \textsc{Hopper} & \textsc{SCon} & \textsc{Tf8} & \textsc{Tf10} \\
\midrule
\makecell{\hspace{-7mm}\textbf{\ourmethod} \\ \hspace{-4mm}(no-regularizer)} & 0.924 & 0.945 & 1.172 & \textbf{0.739} & 0.941 &  \textbf{0.954}  \\
\makecell{\hspace{-7mm}\textbf {\ourmethod} \\ \hspace{-2mm}(with-regularizer)} & \textbf{0.931} & \textbf{0.957} & \textbf{1.572} & 0.732  & \textbf{0.977} & 0.924\\
\bottomrule
\end{tabular}
\caption{Performance comparison between versions of \ourmethod~ with and without regression regularizer at the $100^{\text{th}}$ performance percentile (i.e., maximum solution) generated by each method.} 
\label{tab:3}
\end{table*}

\begin{table*}[ht]
\centering
\begin{tabular}{llllllr} 
\toprule
\hspace{-2mm}\textsc{Method} & \textsc{Ant} & \textsc{DKitty} & \textsc{Hopper} & \textsc{SCon} & \textsc{Tf8} & \textsc{Tf10}  \\
\midrule
\makecell{\hspace{-7mm}\textbf{\ourmethod} \\\hspace{-4mm}(no-regularizer)} & 0.572 & 0.876 & 0.372 & \textbf{0.471} & 0.551 &  \textbf{0.768}  \\
\makecell{\hspace{-7mm}\textbf \ourmethod \\ \hspace{-2mm}(with-regularizer)} & \textbf{0.611} & \textbf{0.887} & \textbf{0.393} & 0.439  & \textbf{0.594} & 0.720\\
\bottomrule
\end{tabular}
\caption{Performance comparison between versions of \ourmethod~ with and without regression regularizer at the $50^{\text{th}}$ performance percentile (i.e., maximum solution) generated by each method.} 
\label{tab:4}
\end{table*}

\label{sec:exp}

\vspace{-1.0ex}

\section{Related Work}
\label{sec:related}
\noindent Black-box optimization problems were studied using derivative-free methods, such as random gradient estimation~\citep{WangAISTATS18} or Bayesian optimization~\citep{Snoek12,Wang13,eriksson2019scalable}. These methods require online evaluation of the target function to approximate its derivative or learn its surrogate model. In many practical applications, this can be very expensive (e.g., testing new protein or drug design), or even dangerous (e.g., test-driving autonomous vehicles in a real physical environment). To avoid this, \textit{offline optimization} approaches tackle this problem via utilizing an existing dataset that records target function evaluations for a fixed set of inputs. These approaches can be categorized into two main families:

\noindent {\bf Conditioning Search Model.}
Existing approaches in this direction are grounded in the framework of density estimation, which aims to learn a probabilistic prior over the input space. The search model is treated as a probability distribution conditioned on the rare event of achieving a high target function score, and is estimated using different approaches, such as adaptive trust-region based strategies~\citep{BrookeICML19}, adaptive step-size in gradient update via reinforcement learning~\citep{YassineAAAI24} or zero-sum game~\citep{ClaraNeurIPs20}, or autoregressive modeling \cite{krishnamoorthy2023generative}, \cite{krishnamoorthy2023diffusion}. \citet{kumar2020model}~learns an inverse mapping of the target function evaluations to inputs and uses it as a search model that predicts which regions will most likely have high-performing designs.
These approaches are often sensitive to the accuracy of the conditioning at out-of-distribution input regimes and/or require learning a computationally expensive generative model of the input space. The robustness of these conditioning algorithms has neither been defined, nor investigated.

\noindent {\bf Conditioning Surrogate Model.} Approaches in this direction tend to fix the search methodology and focus on conditioning the surrogate model to improve the likelihood of finding a good design. This is generally achieved via adopting different forms of regularization on the predicted values of OOD inputs based on the learned surrogate. For example, \citet{YuNeurIPs21}~uses robust model pre-training and adaptation to ensure local smoothness, whereas \citet{FuICLR21}~maximizes data likelihood to reduce the uncertainty in OOD prediction. Alternatively, \citet{TrabuccoICML21}~penalizes high-value predictions for OOD examples, and \citet{NghiaICML24}~penalizes surrogate candidates with high prediction sensitivity over the offline data to avoid overestimation. These approaches are only justified empirically through practical demonstrations. From a theoretical perspective, the extent of effectiveness of these algorithms, as well as the fundamental question regarding when to trust a surrogate function both remain unclear. 

\vspace{-1.0ex}

\section{Conclusion}
\label{sec:conclude}
This paper presents a new theoretical perspective on offline black-box optimization which established the first upper bound on the performance gap between the solutions guided by a trained surrogate and the target function. The bound reveals that such performance gap depends on how well the surrogate model matches the gradient field of the target function on the offline dataset. Inspired by this theory, we studied a novel algorithm for creating surrogate models based on gradient matching and demonstrated improved solutions on diverse real-world benchmarks. Although our theory and algorithm is grounded in the context of offline optimization, the developed principles can be broadly applied to related sub-areas including safe Bayesian optimization and safe reinforcement learning in online learning scenarios.
\section*{Impact Statement}
\label{sec:impact}
This paper introduces a new theoretical perspective to understand and analyze the offline optimization problem, which is a cost-effective alternative to the traditional online experimentation approach to material or experimental design. The methodological improvements and new understanding gained in the paper can lead to improvements in many science and engineering applications including design optimization of hardware, materials, and molecules. Our empirical studies only use publicly available dataset. We do not anticipate any negative ethical or societal impact.

\bibliography{icml2024}
\bibliographystyle{icml2024}

\newpage
\appendix
\onecolumn
\section{Proof of Theorem~\ref{thm:1}}
\label{app:a}
\setcounter{theorem}{0}
\begin{theorem}[Worst-case optimization risk in terms of gradient estimation error] Suppose $g(\mathbf{x})$ is a continuous function with Lipschitz and smooth constants, $\ell$ and $\mu$. Then, we have
\begin{eqnarray}
\mathfrak{G}_{m,\lambda} \ \ \ \triangleq \ \ \ \max_{\mathbf{x}}\mathfrak{G}_{m,\lambda}\Big(\mathbf{x}\Big) &\leq& m\lambda\ell\Big(1 + \lambda\mu\Big)^{m - 1} \cdot \max_{\mathbf{x}}\Big\|\nabla g(\mathbf{x}) - \nabla g_\phi(\mathbf{x})\Big\| \nonumber    
\end{eqnarray}
which characterizes the upper-bound of the worst-case performance gap in terms of the maximum norm difference between the surrogate and oracle gradient over the input space.
\end{theorem}

\begin{proof}
We first note that the performance of the $m$-step oracle solution starting at $\mathbf{x}_\ast^0 = \mathbf{x}^0$ is exactly the ($m$-$1$)-step oracle solution starting at $\mathbf{x}_{\ast}^1$. Likewise, the performance of the $m$-step surrogate solution starting at $\mathbf{x}_\phi^0 =\mathbf{x}^0$ is exactly the ($m$-$1$)-step surrogate solution starting at $\mathbf{x}_{\phi}^1$. That is:
\begin{eqnarray}
    \mathfrak{R}^m_g\left(\mathbf{x}^0\right) \ =\ \mathfrak{R}^m_g\left(\mathbf{x}^0_\ast\right) &=& \mathfrak{R}^{m-1}_g\left(\mathbf{x}_{\ast}^1\right) \quad\text{where}\quad \mathbf{x}_\ast^1 \ =\ \mathbf{x}^0_\ast \ +\ \lambda\nabla g\left(\mathbf{x}^0_\ast\right)\nonumber \\ \mathfrak{R}^m_g\left(\mathbf{x}^0\right) \ =\ \mathfrak{R}^m_{g_\phi}\left(\mathbf{x}^0_\phi\right) &=& \mathfrak{R}^{m-1}_{g_\phi}\left(\mathbf{x}_{\phi}^1\right)\quad\text{where}\quad \mathbf{x}_\phi^1 \ =\ \mathbf{x}^0_\phi \ +\ \lambda\nabla g_\phi\left(\mathbf{x}^0_\phi\right)
\end{eqnarray}
Consequently, for each initial point $\mathbf{x}^0$, we can bound the performance gap as follows:
\begin{eqnarray}
\mathfrak{G}_{m,\lambda}\Big(\mathbf{x}^0\Big)
    &\triangleq& 
    \left\| 
    \mathfrak{R}^m_{g_{\phi}}\left(\mathbf{x}^0\right) \ -\  \mathfrak{R}^m_{g}\left(\mathbf{x}^0\right) 
    \right\| \ =\  
    \left\| 
    \mathfrak{R}^{m-1}_{g_{\phi}}\left(\mathbf{x}^1_{\phi}\right) - 
    \mathfrak{R}^{m-1}_{g}\left(\mathbf{x}^1_{\ast}\right) 
    \right\| \nonumber \\
    &=& 
    \left\| 
    \mathfrak{R}^{m-1}_{g_\phi}\left(\mathbf{x}^1_{\phi}\right) \ -\ 
    \mathfrak{R}^{m-1}_{g}\left(\mathbf{x}^1_{\phi}\right) \
     +\ 
    \mathfrak{R}^{m-1}_{g}\left(\mathbf{x}^1_{\phi}\right) \ -\ 
    \mathfrak{R}^{m-1}_{g}\left(\mathbf{x}^1_{\ast}\right) 
    \right\| \nonumber \\
    &\leq& 
    \left\| 
    \mathfrak{R}^{m-1}_{g_{\phi}}\left(\mathbf{x}^1_{\phi}\right) \ -\  
    \mathfrak{R}^{m-1}_{g}\left(\mathbf{x}^1_{\phi}\right)
    \right\|
    \ +\  
    \Big\|
    \mathfrak{R}^{m-1}_{g}\left(\mathbf{x}^1_{\phi}\right) \ -\  
    \mathfrak{R}^{m-1}_{g}\left(\mathbf{x}^1_{\ast}\right) 
    \Big\| 
    \nonumber\\
    &=& \mathfrak{G}_{m-1, \lambda}\left(\mathbf{x}_\phi^1\right) \ +\ \Big\|
    \mathfrak{R}^{m-1}_{g}\left(\mathbf{x}^1_{\phi}\right) \ -\  
    \mathfrak{R}^{m-1}_{g}\left(\mathbf{x}^1_{\ast}\right) 
    \Big\| \ .
\label{thm1-e1}
\end{eqnarray}
Thus, let $\mathcal{E}_{m-1}(\mathbf{x}_\phi^1, \mathbf{x}_\ast^1) \triangleq \|
    \mathfrak{R}^{m-1}_{g}(\mathbf{x}^1_{\phi}) \ -\  
    \mathfrak{R}^{m-1}_{g}(\mathbf{x}^1_{\ast}) 
    \|$, we have $\mathfrak{G}_{m,\lambda}(\mathbf{x}^0) \leq \mathfrak{G}_{m - 1,\lambda}(\mathbf{x}_\phi^1) + \mathcal{E}_{m-1}(\mathbf{x}_\phi^1, \mathbf{x}_\ast^1)$. To bound the term $\mathcal{E}_{m-1}(\mathbf{x}_\phi^1, \mathbf{x}_\ast^1)$, we will prove the following intermediate results. 

\begin{lemma}[]
For any $k \in [1, m]$ and two different starting points $\mathbf{u}^0$ and $\mathbf{v}^0$, the performance gap between the $k$-step oracle solutions respectively starting from $\mathbf{u}^0$ and $\mathbf{v}^0$ is bounded by the norm distance between the starting points:
\begin{eqnarray}
\mathcal{E}_k\Big(\mathbf{u}^0, \mathbf{v}^0\Big) \ \ =\ \ \left\|
\mathfrak{R}^{k}_{g}\left(\mathbf{u}^0\right) \ -\  
\mathfrak{R}^{k}_{g}\left(\mathbf{v}^0\right) 
\right\| 
&\leq& 
\ell(1 + \lambda\mu)^{k} \left\|
\mathbf{u}^{0} \ -\  
\mathbf{v}^{0}
\right\| \ .
\end{eqnarray}
\label{lem:1}
\end{lemma}
\begin{proof}
Let us first define the respective oracle search trajectories using the same gradient ascent formalism. That is, the respective candidate solutions at some intermediate step $\kappa \in [1, k]$ are given by $\mathbf{u}^\kappa = \mathbf{u}^{\kappa-1} + \lambda \nabla g(\mathbf{u}^{\kappa-1})$ and $\mathbf{v}^\kappa = \mathbf{v}^{\kappa-1} + \lambda \nabla g(\mathbf{v}^{\kappa-1})$. We can then make use of the Lipschitz continuous assumption to achieve the following bound:
\begin{eqnarray}
\left\|
\mathfrak{R}^{k}_{g}\left(\mathbf{u}^0\right) - 
\mathfrak{R}^{k}_{g}\left(\mathbf{v}^0\right) 
\right\| 
&=&
\left\|
g\left(\mathbf{x}_\ast\right) - g\left(\mathbf{u}^k\right)
- g\left(\mathbf{x}_\ast\right) \ +\  g\left(\mathbf{v}^k\right)
\right\| \nonumber \\ 
&=&
\left\|
g\left(\mathbf{v}^k\right)
- g\left(\mathbf{u}^k\right)
\right\| \ \ \leq\  \ 
\ell \cdot \left\|
\mathbf{v}^k
- \mathbf{u}^k
\right\| \ .
\label{lem1-e1}
\end{eqnarray}
We subsequently bound the distance between the candidate solutions at step $k$ in terms of the distance at step $k-1$ using the smoothness conditions:
\begin{eqnarray}
\left\| \mathbf{v}^{k} - \mathbf{u}^{k} \right\| 
&=&
\left\|
\mathbf{v}^{k-1} - 
\mathbf{u}^{k-1} + 
\lambda \nabla g\left(\mathbf{v}^{k-1}\right) \ -\ 
\lambda \nabla g\left(\mathbf{u}^{k-1}\right)
\right\| \nonumber \\
&\leq&
\left\|
\mathbf{v}^{k-1} - 
\mathbf{u}^{k-1}
\right\| 
\ +\  
\lambda \Big\|
\nabla g\left(\mathbf{v}^{k-1}\right) \ - \ 
\nabla g\left(\mathbf{u}^{k-1}\right)
\Big\| 
\nonumber \\
&\leq& 
(1 + \lambda\mu) \left\|
\mathbf{v}^{k-1} - 
\mathbf{u}^{k-1}
\right\| \ .
\end{eqnarray}
Applying this bound recursively yields
$\| \mathbf{v}^{k} - \mathbf{u}^{k} \| 
\leq
(1 + \lambda\mu)^{k} \|
\mathbf{v}^{0} - 
\mathbf{u}^{0}\|$. We finally substitute the above into ~\eqref{lem1-e1} to arrive at the final bound $\|
\mathfrak{R}^{k}_{g}(\mathbf{u}^0) - 
\mathfrak{R}^{k}_{g}(\mathbf{v}^0) 
\| \leq \ell(1 + \lambda\mu)^{k} \|
\mathbf{v}^{0} - 
\mathbf{u}^{0}
\|$.
\end{proof}

Applying Lemma~\ref{lem:1} with $k=m-1$, $\mathbf{u}^0 = \mathbf{x}^1_\phi$, and $\mathbf{v}^0 = \mathbf{x}^1_\ast$ subsequently allows us to bound $\mathcal{E}_{m-1}(\mathbf{x}_\phi^1, \mathbf{x}_\ast^1)$ as follows:
\begin{eqnarray}
\mathcal{E}_{m-1}\left(\mathbf{x}_\phi^1, \mathbf{x}_\ast^1\right) 
&\leq& 
\ell(1 + \lambda\mu)^{m-1}\left\|\mathbf{x}^1_\phi - \mathbf{x}^1_\ast\right\| \nonumber \\
&=& 
\ell(1 + \lambda\mu)^{m-1}\left\|\mathbf{x}^0 + \lambda\nabla g_\phi(\mathbf{x}^0) - \mathbf{x}^0 - \lambda\nabla g(\mathbf{x}^0)\right\| \nonumber \\
&=&
\lambda\ell(1 + \lambda\mu)^{m-1}\left\|\nabla g_\phi(\mathbf{x}^0) - \nabla g(\mathbf{x}^0)\right\| \ \ \triangleq \  \ \mathcal{Q}\left(\mathbf{x}^0\right) \ . \label{eq:a}
\end{eqnarray}

We will now use this result to complete our bound for the performance gap. That is:
\begin{eqnarray}
\mathfrak{G}_{m,\lambda} 
&\triangleq& \underset{\mathbf{x}^0}{\mathrm{max}} \ \mathfrak{G}_{m,\lambda}\Big(\mathbf{x}^0\Big)
\ \leq \ 
\underset{\mathbf{x}^0}{\mathrm{max}} \ \mathfrak{G}_{m-1,\lambda}\Big(\mathbf{x}^1_\phi\Big) \ +\  
\underset{\mathbf{x}^0}{\mathrm{max}} \ \mathcal{Q}\Big(\mathbf{x}^0\Big) \nonumber \\
&\leq& 
\mathfrak{G}_{m-1,\lambda} \ \ +\ \ 
\underset{\mathbf{x}^0}{\mathrm{max}} \ \mathcal{Q}\Big(\mathbf{x}^0\Big) \nonumber \\
&\leq& 
\mathfrak{G}_{0,\lambda} \ \ +\ \ 
m \cdot \underset{\mathbf{x}^0}{\mathrm{max}} \ \mathcal{Q}\Big(\mathbf{x}^0\Big) \ ,
\label{thm1-e2}
\end{eqnarray}
where the last inequality is obtained via recursively applying the previous inequality $m$ times. Substituting $\mathfrak{G}_{0,\lambda}\ =\ 0$ and the upper-bound for $\mathcal{Q}\Big(\mathbf{x}^0\Big)$ above into ~\eqref{thm1-e2} gives:
\begin{eqnarray}
\mathfrak{G}_{m,\lambda} &\leq& m\lambda\ell(1 + \lambda\mu)^{m-1} \ \cdot\ 
\underset{\mathbf{x}^0}{\mathrm{max}} \ 
\Big\|
\nabla g_\phi\left(\mathbf{x}^0\right) - 
\nabla g\left(\mathbf{x}^0\right)
\Big\| \ ,
\end{eqnarray}
which completes our proof for Theorem~\ref{thm:1}.
\end{proof}

\noindent {\bf Tightness of the bound.} Note that despite the exponential dependence on $m$ of the above bound, its tightness can be controlled by choosing a sufficiently small value for $\lambda$. For example, if we choose $\lambda \leq 1/m$, it will follow that 
\begin{eqnarray}
\left(1 + \lambda \cdot \mu\right)^{m-1} &\leq& \left(1 + \frac{\mu}{m}\right)^{m-1} \ \ <\ \ \left(1 + \frac{\mu}{m}\right)^m
\end{eqnarray}
which will approach $e^\mu$ in the limit of $m$. Here, we use the known fact that $\mathrm{lim}_{m\rightarrow\infty}(1 + \mu/m)^m = e^\mu$ with $\mu > 0$. As such, when $m$ is sufficiently large the bound in Theorem~\ref{thm:1} is upper-bounded with $m \cdot\lambda \cdot \ell \cdot (1 + \lambda \cdot \mu)^{m - 1} \cdot \text{gradient-gap} \simeq \ell \cdot e^\mu \cdot\text{gradient-gap}$ which asserts that the worst-case performance gap of our offline optimizer is approaching (in the limit of $m$) $\ell \cdot e^\mu \cdot \max_{\mathbf{x}}\|\nabla g(\mathbf{x}) - \nabla g_\phi(\mathbf{x})\| = \mathbf{O}(\max_{\mathbf{x}}\|\nabla g(\mathbf{x}) - \nabla g_\phi(\mathbf{x})\|)$ which is not dependent on the no. of gradient steps.

\section{Minimizing Eq.~\eqref{eq:11} Reduces Gradient Gap.}
\label{app:gradient-gap}
Intuitively, minimizing Eq.~\ref{eq:11} will reduce the gradient gap. To formalize this intuition rigorously, we will show below that (1) in the limit of optimization if a parameterization $\phi$ can be found that zeroes out the loss in Eq.~\ref{eq:11} over the entire input space, the gradient gap is zero; and (2) in more practical cases, where the loss in Eq.~\ref{eq:11} is not zero, the gradient gap is still guaranteed to be upper-bound by the Lipschitz constant of the function gap, which decreases as we optimize the loss function in Eq.~\ref{eq:11}. These are detailed below.\\

\noindent {\bf A. The minimized loss in Eq.~\ref{eq:11} is zero.} In this case, let us define:
\begin{eqnarray}
\mathbf{F}_{\phi}(\mathbf{x}, \mathbf{x}') &\triangleq& \int_0^1 \nabla g(t\mathbf{x} + (1 - t)\mathbf{x}')\mathrm{d}t - \int_0^1 \nabla g_\phi(t\mathbf{x} + (1 - t)\mathbf{x}')\mathrm{d}t \ .
\end{eqnarray}
The loss in Eq.~\ref{eq:11} can be rewritten as 
\begin{eqnarray}
\mathfrak{L}_g(\phi) &=& \mathbb{E}\left[\Bigg(\mathbf{F}_\phi\Big(\mathbf{x},\mathbf{x}'\Big)^\top\Big(\mathbf{x} - \mathbf{x}'\Big)\Bigg)^2\right] \ ,
\end{eqnarray}
where the expectation is over all pairs $(\mathbf{x},\mathbf{x}')$ from the input space. At the optimal $\phi$, since $\mathfrak{L}_g(\phi) = 0$ as assumed, 
\begin{eqnarray}
\mathbf{F}_\phi\Big(\mathbf{x},\mathbf{x}'\Big)^\top\Big(\mathbf{x} - \mathbf{x}'\Big) &=& 0 \label{eq:d1}
\end{eqnarray}
for any choice of $(\mathbf{x},\mathbf{x}')$. Next, by the line integration theorem, we also have
\begin{eqnarray}
g(\mathbf{x}) - g(\mathbf{x}') &=& (\mathbf{x} - \mathbf{x}')^\top\left(\int_0^1 \nabla g(t\mathbf{x} + (1 - t)\mathbf{x}')\mathrm{d}t\right) \ , \\
g_\phi(\mathbf{x}) - g_\phi(\mathbf{x}') &=& (\mathbf{x} - \mathbf{x}')^\top\left(\int_0^1 \nabla g_\phi(t\mathbf{x} + (1 - t)\mathbf{x}')\mathrm{d}t\right) \ ,
\end{eqnarray}
which together imply that
\begin{eqnarray}
\mathbf{F}_\phi\Big(\mathbf{x},\mathbf{x}'\Big)^\top\Big(\mathbf{x} - \mathbf{x}'\Big) &=& g(\mathbf{x}) - g(\mathbf{x}') - g_\phi(\mathbf{x}) + g_\phi(\mathbf{x}') \ .\label{eq:d2}
\end{eqnarray}
Combining Eq.~\ref{eq:d1} and Eq.~\ref{eq:d2} results in
\begin{eqnarray}
g(\mathbf{x}) - g_\phi(\mathbf{x}) &=& g(\mathbf{x}') - g_\phi(\mathbf{x}') \label{eq:d3}
\end{eqnarray}
for any choice of $(\mathbf{x}, \mathbf{x}')$. This means there exists a constant $c$ such that
\begin{eqnarray}
g(\mathbf{x}) - g_\phi(\mathbf{x}) &=& c
 \label{eq:d4}
\end{eqnarray}
for all $\mathbf{x}$. Thus, taking the derivative with respect to $\mathbf{x}$ on both sides of the above yields
\begin{eqnarray}
\nabla g(\mathbf{x}) - \nabla g_\phi(\mathbf{x}) &=& 0 \ ,
 \label{eq:d5}
\end{eqnarray}
which implies immediately that the gradient gap is zero everywhere. Hence, optimizing Eq.~\ref{eq:11} guarantees in principle that the gradient will be perfectly matched in the limit of data (i.e., when we take the expectation over the entire input space rather than over a finite set of offline data points).\\

\noindent {\bf B. The minimized loss in Eq.~\ref{eq:11} is not zero.} In this case, let us define
\begin{eqnarray}
h(\mathbf{x}) &=& g(\mathbf{x}) - g_\phi(\mathbf{x}) 
 \label{eq:d6}
\end{eqnarray}
and it will follow that $|\mathbf{F}_\phi(\mathbf{x}, \mathbf{x}')^\top(\mathbf{x} - \mathbf{x}')| = |h(\mathbf{x}) - h(\mathbf{x}')|$ following Eq.~\ref{eq:d2} above. This means our loss function is working towards minimizing $(h(\mathbf{x}) - h(\mathbf{x}'))^2$ over $(\mathbf{x}, \mathbf{x}')$. This will makes $h(\mathbf{x})$ smoother as the output distance between different inputs are being reduced.

As a result, this process will reduce the Lipschitz constant $\epsilon$ of $h(\mathbf{x})$, which is defined to be the minimum value such that
\begin{eqnarray}
\left|h(\mathbf{x}) - h(\mathbf{x}')\right| &\leq& \epsilon \cdot \left\|\mathbf{x} - \mathbf{x}'\right\| \ ,
\end{eqnarray}
which implies
\begin{eqnarray}
\left|(g(\mathbf{x}) - g(\mathbf{x}')) - (g_\phi(\mathbf{x}) - g_\phi(\mathbf{x}'))\right| &\leq& \epsilon \cdot \left\|\mathbf{x} - \mathbf{x}'\right\|
\end{eqnarray}
Now, dividing both sides by $\|\mathbf{x} - \mathbf{x}'\|$ yields
\begin{eqnarray}
\left|\frac{g(\mathbf{x}) - g(\mathbf{x}')}{\|\mathbf{x} - \mathbf{x}'\|} - \frac{g_\phi(\mathbf{x}) - g_\phi(\mathbf{x}')}{\|\mathbf{x} - \mathbf{x}'\|}\right| &\leq& \epsilon \ .
\end{eqnarray}
Now, suppose we choose $\mathbf{x}' = \mathbf{x} + t \cdot \mathbf{e}_i$ where $\mathbf{e}_i$ is a $d$-dimensional one-hot vector with the hot component at the $i$-th position where $d$ denotes the input dimension. So, the above is equivalent to
\begin{eqnarray}
\left|\frac{g(\mathbf{x} + t \cdot \mathbf{e}_i) - g(\mathbf{x})}{t\|\mathbf{e}_i\|} - \frac{g_\phi(\mathbf{x} + t \cdot \mathbf{e}_i) - g_\phi(\mathbf{x})}{t\|\mathbf{e}_i\|}\right| &\leq& \epsilon \ ,
\end{eqnarray}
or more expressively,
\begin{eqnarray}
- \epsilon \ \ \leq \ \ \frac{g(\mathbf{x} + t \cdot \mathbf{e}_i) - g(\mathbf{x})}{t\|\mathbf{e}_i\|} - \frac{g_\phi(\mathbf{x} + t \cdot \mathbf{e}_i) - g_\phi(\mathbf{x})}{t\|\mathbf{e}_i\|} \ \ \leq \ \ \epsilon 
\end{eqnarray}
Taking $\mathrm{lim}_{t \rightarrow 0}$ on all parts of the above inequality, the above can be rewritten as
\begin{eqnarray}
-\epsilon \ \ \leq \ \ \mathrm{lim}_{t\rightarrow 0}\left(\frac{g(\mathbf{x} + t \cdot \mathbf{e}_i) - g(\mathbf{x})}{t\|\mathbf{e}_i\|}\right) - \mathrm{lim}_{t\rightarrow 0}\left(\frac{g_\phi(\mathbf{x} + t \cdot \mathbf{e}_i) - g_\phi(\mathbf{x})}{t\|\mathbf{e}_i\|}\right) \ \ \leq \ \ \epsilon \ . 
\end{eqnarray}
Next, using the definition of directional gradient
\begin{eqnarray}
\nabla_{\mathbf{r}} \ g(\mathbf{x})  &=& \mathrm{lim}_{t\rightarrow 0}\ \ \frac{1}{t}\Bigg(g(\mathbf{x} + t \cdot\mathbf{r}) - g(\mathbf{x})\Bigg)
\end{eqnarray}
and the fact that $\nabla_{\mathbf{r}} g(\mathbf{x}) = \nabla g(\mathbf{x})^\top\mathbf{r}$ on $\mathbf{r} = \mathbf{e}_i$, we have
\begin{eqnarray}
-\epsilon \ \ \leq \ \ \frac{1}{\|\mathbf{e}_i\|} \cdot \Big(\nabla g(\mathbf{x})^\top \mathbf{e}_i\Big) - \frac{1}{\|\mathbf{e}_i\|} \cdot \Big(\nabla g_\phi(\mathbf{x})^\top \mathbf{e}_i\Big) \ \ \leq \ \ \epsilon \ ,
\end{eqnarray}
which implies that
\begin{eqnarray}
\Big(\nabla g(\mathbf{x}) - \nabla g_\phi(\mathbf{x}) \Big)^\top\mathbf{e}_i &\leq& \epsilon \cdot \|\mathbf{e}_i\| \ \ =\ \ \epsilon \ .
\end{eqnarray}
The last step is true because $\|\mathbf{e}_i\| = 1$. Next, repeat the above argument with $\mathbf{r} = \mathbf{e}_i$ for $i = 1, 2, \ldots, d$ and summing both sides of the resulting inequalities over $i = 1, 2, \ldots, d$, we have
\begin{eqnarray}
\Big\|\nabla g(\mathbf{x}) - \nabla g_\phi(\mathbf{x})\Big\|_1 \ \ \triangleq\ \ \sum_{i=1}^d \Bigg[\Big(\nabla g(\mathbf{x}) - \nabla g_\phi(\mathbf{x}) \Big)^\top\mathbf{e}_i\Bigg] &\leq& d \cdot \epsilon \ =\ \mathbf{O}(\epsilon) \ .
\end{eqnarray}
Finally, we note that
\begin{eqnarray}
\Big\|\nabla g(\mathbf{x}) - \nabla g_\phi(\mathbf{x})\Big\|_2 \ \ \leq \ \ \Big\|\nabla g(\mathbf{x}) - \nabla g_\phi(\mathbf{x})\Big\|_1 &\leq& \mathbf{O}(\epsilon) \ ,
\end{eqnarray}
which completes our proof and asserts that the gradient gap is indeed bounded by the Lipschitz constant of the gap function $h(\mathbf{x}) = g(\mathbf{x}) - g_\phi(\mathbf{x})$ that decreases as we optimize the training objective.
\section{Training and Evaluation Details of \ourmethod} \label{appendix:training_eval}
We used a feed-forward neural network with $4$ layers ($512 \rightarrow 128 \rightarrow 32 \rightarrow 1$) activated by the Leaky ReLU function as the surrogate model for \ourmethod. For each task, we trained the model using Adam optimizer with 1e-4 learning rate and a batch size of 128 for 200 epochs.   

During the evaluation, we employed gradient updates for 150 iterations uniformly across all the tasks. This evaluation procedure used an Adam optimizer with a 0.01 learning rate for all discrete tasks, and a 0.001 learning rate for all continuous tasks. We chose a larger learning rate for discrete tasks since the discrete inputs are converted into logits (same as all baselines). 

\section{Extended Theoretical Analysis to Incorporate the Value Matching Regularizer}
\label{app:regularizer-analysis}
This section discusses an extension of the original theoretical analysis in Section~\ref{sec:theory} to provide a theoretical condition under which the worst-case optimization in Theorem~\ref{thm:1} has a tighter bound. We will show that this bound is expressed in terms of both the gradient and value matching quantities, which inspires the addition of the value-matching regularizer in Eq.~\eqref{eq:14} to the original loss in Eq.~\eqref{eq:13}. This is formalized in Theorem~\ref{thm:2} below.

\begin{theorem}[Generalized worst-case optimization risk bound]
\label{thm:2}
Suppose the target objective function $g(\mathbf{x})$ is a $\ell$-Lipschitz continuous and $\mu$-Lipschitz smooth function. For all $a \in (0, 1)$, the worst-case performance gap, $\mathfrak{G}_{m, \lambda} \triangleq \max_{\mathbf{x}}\mathfrak{G}_{m,\lambda}(\mathbf{x})$, between $g$ and some arbitrary surrogate $g_{\phi}$ with Lipschitz constant $\ell_\phi$ is upper-bounded by:
\begin{eqnarray}
\mathfrak{G}_{m, \lambda} &\leq& m \cdot 2a \cdot \max_{\mathbf{x}} \Big\|g(\mathbf{x}) - g_\phi(\mathbf{x})\Big\| \nonumber\\
&+& m \cdot \Big(\ell + a \cdot (\ell_\phi - \ell)\Big) \cdot \Big(1 + \lambda \mu\Big)^{m-1} \cdot \max_{\mathbf{x}} \Big\|\nabla_{\mathbf{x}}g(\mathbf{x}) - \nabla_{\mathbf{x}}g_\phi(\mathbf{x})\Big\| \ .\label{eq:r1}
\end{eqnarray}
\end{theorem}

\begin{proof}
First, let us start from Eq.~\eqref{lem1-e1}, 
\begin{eqnarray}
\Big\| \mathfrak{R}^k_g\big(\mathbf{u}^0\big) - \mathfrak{R}^k_g\big(\mathbf{v}^0\big) \Big\| \ \ =\ \  \Big\| g\big(\mathbf{v}^k\big) - g\big(\mathbf{u}^k\big) \Big\| &=& \Big\| g\big(\mathbf{v}^k\big) - g_{\phi} \big(\mathbf{v}^k\big) + g_{\phi}\big(\mathbf{v}^k\big) - g_{\phi}\big(\mathbf{u}^k\big) + g_{\phi}\big(\mathbf{u}^k\big) - g\big(\mathbf{u}^k\big) \Big\| \nonumber\\
&\leq& \Big\| g\big(\mathbf{v}^k\big) - g_{\phi} \big(\mathbf{v}^k\big) \Big\| + \Big\| g_{\phi}\big(\mathbf{v}^k\big) - g_{\phi}\big(\mathbf{u}^k\big) \Big\| + \Big\| g_{\phi}\big(\mathbf{u}^k\big) - g\big(\mathbf{u}^k\big) \Big\| \nonumber\\
\label{eq:r2}
\end{eqnarray}
Next, similar to the derivation of Lemma~\ref{lem:1}, we have:
\begin{eqnarray}
\Big\| g_{\phi}\big(\mathbf{v}^k\big) - g_{\phi}\big(\mathbf{u}^k\big) \Big\| &\leq& \ell_\phi \cdot (1 + \lambda \mu)^k \cdot \Big\| \mathbf{v}^0 - \mathbf{u}^0 \Big\| \label{eq:r3}
\end{eqnarray}
with $\ell_\phi$ denotes the Lipschitz constant of the surrogate $g_\phi(\mathbf{x})$.

Now, let $\mathcal{E}_k(\mathbf{u}^0, \mathbf{v}^0) \triangleq |\mathfrak{R}_g^k(\mathbf{u}^0) - \mathfrak{R}_g^k(\mathbf{v}^0)|$ as defined in Lemma~\ref{lem:1} of Appendix~\ref{app:a}. It follows that
\begin{eqnarray}
\mathcal{E}_{k}\big(\mathbf{u}^0, \mathbf{v}^0\big) &\leq& \Big\|g\big(\mathbf{v}^k\big) - g_{\phi} \big(\mathbf{v}^k\big)\Big\| \ +\  \Big\|g_{\phi}\big(\mathbf{u}^k\big) - g\big(\mathbf{u}^k\big)\Big\| \ +\  \ell_\phi \cdot (1 + \lambda \mu)^k \cdot \Big\|\mathbf{v}^0 - \mathbf{u}^0\Big\| \ .\label{eq:r4}
\end{eqnarray}
Let $k=m-1$, $\mathbf{u}^0 = \mathbf{x}^1_\phi$, $\mathbf{v}^0 = \mathbf{x}^1_\ast$. Similar to Eq.~\eqref{eq:a}, we have
\begin{eqnarray}
\mathcal{E}_{m-1}\big(\mathbf{x}^1_\phi, \mathbf{x}^1_\ast\big) &\leq& \Big\|g(\mathbf{v}^{m-1}) - g_{\phi} (\mathbf{v}^{m-1})\Big\| + \Big\|g_{\phi}(\mathbf{u}^{m-1}) - g(\mathbf{u}^{m-1})\Big\| \nonumber\\
&+& \ell_\phi \cdot (1 + \lambda \mu)^{m-1} \cdot \Big\|
\nabla_\mathbf{x}g\big(\mathbf{x}^0\big) - \nabla_\mathbf{x}g(\mathbf{x}^0\big)\Big\| \ . \label{eq:r5}
\end{eqnarray}
Additionally, the following was also shown in Lemma~\ref{lem:1}:
\begin{eqnarray}
\mathcal{E}_{m-1}\big(\mathbf{x}^1_\phi, \mathbf{x}^1_\ast\big) &\leq& \lambda\ell \cdot (1 + \lambda \mu)^{m-1} \cdot \Big\|\nabla_{\mathbf{x}}g\big(\mathbf{x}^0\big) - \nabla_{\mathbf{x}}g_\phi\big(\mathbf{x}^0\big)\Big\| \ . \label{eq:r6}
\end{eqnarray}
Now, combining Eq.~\eqref{eq:r5} and Eq.~\eqref{eq:r6} with $a \in (0, 1)$,
\begin{eqnarray}
\mathcal{E}_{m-1}\big(\mathbf{x}^1_\phi, \mathbf{x}^1_\ast\big) &\leq& a \cdot \Bigg(\Big\|g\big(\mathbf{v}^{m-1}\big) - g_{\phi} \big(\mathbf{v}^{m-1}\big)\Big\| \ +\  \Big\|g_{\phi}\big(\mathbf{u}^{m-1}\big) - g\big(\mathbf{u}^{m-1}\big)\Big\|\Bigg) \nonumber\\
&+& \Big((1-a)\cdot \lambda\ell + a \cdot \ell_\phi)\Big) \cdot (1 + \lambda \mu)^{m-1} \cdot \Big\|\nabla_{\mathbf{x}}g\big(\mathbf{x}^0\big) - \nabla_{\mathbf{x}}g_\phi\big(\mathbf{x}^0\big)\Big\| \ .\label{eq:r7}
\end{eqnarray}
Plugging the above new bound to Eq.~\eqref{thm1-e2} in Appendix~\ref{app:a},
\begin{eqnarray}
\mathfrak{G}_{m,\lambda} \ = \ \max_{
\mathbf{x}^0} \ \mathfrak{G}_{m, \lambda} \big(\mathbf{x}^0\big) &\leq& \max_{\mathbf{x}^0} \ \mathfrak{G}_{m-1, \lambda} \big(\mathbf{x}^0\big)  \nonumber \\
&+& \max_{\mathbf{x}^0} \ a \cdot \Bigg(\Big\|g\big(\mathbf{v}^{m-1}\big) - g_{\phi} \big(\mathbf{v}^{m-1}\big)\Big\| \ +\  \Big\|g_{\phi}\big(\mathbf{u}^{m-1}\big) - g\big(\mathbf{u}^{m-1}\big)\Big\|\Bigg) \nonumber\\
&+& \max_{\mathbf{x}^0} \Big((1-a)\cdot \lambda\ell + a \cdot \ell_\phi)\Big) \cdot (1 + \lambda \mu)^{m-1} \cdot \Big\|\nabla_{\mathbf{x}}g\big(\mathbf{x}^0\big) - \nabla_{\mathbf{x}}g_\phi\big(\mathbf{x}^0\big)\Big\| \ .\label{eq:r8}
\end{eqnarray}
Rearranging the above gives the following upper bound:
\begin{eqnarray}
\mathfrak{G}_{m,\lambda} - \mathfrak{G}_{m-1, \lambda} &\leq& \max_{\mathbf{x}^0} \ 2a \cdot \Big\|g\big(\mathbf{x}^0\big) - g_{\phi} \big(\mathbf{x}^0\big)\Big\| \nonumber\\
&+& \max_{\mathbf{x}^0} \ \Big((1-a)\cdot \lambda\ell + a \cdot \ell_\phi)\Big) \cdot \Big(1 + \lambda \mu\Big)^{m-1} \Big\|\nabla_{\mathbf{x}}g\big(\mathbf{x}^0\big) - \nabla_{\mathbf{x}}g_\phi\big(\mathbf{x}^0\big)\Big\| \ ,\label{eq:r9}
\end{eqnarray}
where the first term on the RHS is derived from the fact that the maximum of an unconstrained optimization program upper bounds that of a constrained optimization program (the dependency of $\mathbf{u}^{m-1}$ and $\mathbf{v}^{m-1}$ on $\mathbf{x}^0$ can be seen as constraints). Applying this $m$ times consecutively gives:
\begin{eqnarray}
\mathfrak{G}_{m,\lambda} - \mathfrak{G}_{0, \lambda} &\leq& 2am \cdot \max_{\mathbf{x}} \Big\|g(\mathbf{x}) - g_\phi(\mathbf{x})\Big\| \nonumber\\
&+& m \cdot \Big((1-a)\cdot \lambda\ell + a \cdot \ell_\phi)\Big) \cdot \Big(1 + \lambda \mu\Big)^{m-1} \cdot \max_{\mathbf{x}} \Big\|\nabla_{\mathbf{x}}g\big(\mathbf{x}\big) - \nabla_{\mathbf{x}}g_\phi\big(\mathbf{x}\big)\Big\| \ .
\end{eqnarray}
Since $\mathfrak{G}_{0, \lambda} = 0$, this implies:
\begin{eqnarray}
\mathfrak{G}_{m,\lambda}
&\leq& 2am \cdot \max_{\mathbf{x}} \Big\|g(\mathbf{x}) - g_\phi(\mathbf{x})\Big\| \nonumber\\
&+& m \cdot \Big((1-a)\cdot \lambda\ell + a \cdot \ell_\phi)\Big) \cdot \Big(1 + \lambda \mu\Big)^{m-1} \cdot \max_{\mathbf{x}} \Big\|\nabla_{\mathbf{x}}g\big(\mathbf{x}\big) - \nabla_{\mathbf{x}}g_\phi\big(\mathbf{x}\big)\Big\| \ . \label{eq:r10}
\end{eqnarray}
Using the above result, we can compute the difference between the RHS of this bound in Eq.~\eqref{eq:r10} and that of the original bound in Eq.~\eqref{eq:7}, which is $2am \cdot \max_{\mathbf{x}} \|g(\mathbf{x}) - g_\phi(\mathbf{x})\| + am \cdot (\ell_\phi - \lambda\ell) \cdot (1 + \lambda \mu)^{m-1} \cdot \max_{\mathbf{x}} \|\nabla_\mathbf{x} g(\mathbf{x}) - \nabla_{\mathbf{x}} g_\phi(\mathbf{x})\|$. The second term of this difference could be negative as $\ell_\phi$ decreases beyond $\lambda\ell$ (see Appendix~\ref{app:gradient-gap} for intuition on why $\ell_\phi$ decreases as we train $g_\phi$). As a result, when $\ell_\phi$ and the ratio $\max_{\mathbf{x}} \|g(\mathbf{x}) - g_\phi(\mathbf{x})\|/\max_{\mathbf{x}} \|\nabla_{\mathbf{x}}g\big(\mathbf{x}\big) - \nabla_{\mathbf{x}}g_\phi\big(\mathbf{x}\big)\|$ is sufficiently small, or such that:
\begin{eqnarray}
\ell_\phi \ + \ \frac{ 2 \cdot \max_{\mathbf{x}} \Big\|g(\mathbf{x}) -  g_\phi(\mathbf{x})\Big\| } { \left(1 + \lambda \mu\right)^{m-1} \cdot \max_{\mathbf{x}} \Big\|\nabla_{\mathbf{x}}g\big(\mathbf{x}\big) - \nabla_{\mathbf{x}}g_\phi\big(\mathbf{x}\big)\Big\|}&\leq& \lambda\ell  \ , \label{eq:57}
\end{eqnarray}
the bound will become tighter, thus justifying the importance of the value matching term. Developing a training algorithm to substantiate the above condition will be part of our follow-up work.
\end{proof}

\section{Mean and Standard Deviation Results}
\label{app:stddev}
As mentioned in the evaluation methodology, we ran each method for 4 different runs. This section reports the mean results from Tables~\ref{tab:1} and~\ref{tab:2} along with the corresponding standard deviations. 

\begin{table*}[ht]
\centering
\begin{tabular}{lccc}
\toprule
\textsc{Method} & \textsc{Ant} & \textsc{DKitty} & \textsc{Hopper} \\
\midrule
\textsc{GA} & 0.271 $\pm$ 0.013 & 0.895 $\pm$ 0.013  & 0.780 $\pm$ 0.462 \\
\textsc{Ens-Mean} & 0.517 $\pm$ 0.039 & 0.899 $\pm$ 0.010 & 1.524 $\pm$ 0.710 \\
\textsc{Ens-Min} & 0.536 $\pm$0.031& 0.908  $\pm$0.019& 1.42 $\pm$0.645\\
\textsc{CMA-ES} & \textbf{0.974} $\pm$ 0.556 & 0.722 $\pm$ 0.001 & 0.620 $\pm$0.151 \\
\textsc{MINS} & 0.910 $\pm$ 0.034 & 0.939 $\pm$ 0.003 & 0.150 $\pm$ 0.186 \\
\textsc{CbAS} & 0.842 $\pm$ 0.015 & 0.879 $\pm$ 0.002 & 0.150 $\pm$ 0.014 \\
\textsc{RoMA} & 0.832 $\pm$ 0.055& 0.880 $\pm$ 0.008& 2.026 $\pm$ 0.225 \\
\textsc{BONET} & 0.927 $\pm 0.002 $ & 0.954 $\pm$ 0.0001 & 0.395 $\pm$ 0.0002 \\
\textsc{COMS} & 0.885  $\pm$ 0.024 & 0.953 $\pm$ 0.016 & \textbf{2.270} $\pm 0.237 $ \\
\midrule


\makecell{\textbf {\ourmethod}}& 0.931 $\pm$ 0.011 (2) & \textbf{0.957 $\pm$ 0.014 (1)} & 1.572 $\pm$ 0.322 (3) \\
\midrule
\textsc{Method} & \textsc{SCon} & \textsc{Tf8} & \textsc{Tf10}  \\
\midrule
\textsc{GA} & 0.699 $\pm$ 0.054 & 0.954 $\pm$  0.020 & 0.966 $\pm$ 0.026 \\
\textsc{Ens-Mean} & 0.716 $\pm$ 0.065 & 0.926 $\pm$0.005& \textbf{0.968} $\pm$ 0.019\\
\textsc{Ens-Min} & 0.734 $\pm$ 0.058 & 0.959 $\pm$ 0.052 & 0.959 $\pm$ 0.021 \\
\textsc{CMA-ES} & \textbf{0.757} $\pm$ 0.013 & \textbf{0.978} $\pm$ 0.007 & 0.966 $\pm$ 0.007 \\
\textsc{MINS} & 0.690 $\pm$ 0.024 & 0.900 $\pm$ 0.059 & 0.759 $\pm$0.031 \\
\textsc{CbAS} & 0.659 $\pm$0.086 & 0.916 $\pm$ 0.035& 0.928 $\pm$0.013 \\
\textsc{RoMA} & 0.704 $\pm$0.032 & 0.664  $\pm$0.015 & 0.820 $\pm$0.014 \\
\textsc{BONET} & 0.500 $\pm$0.002 & 0.911 $\pm$ 0.005& 0.756 $\pm 0.006$\\
\textsc{COMS} & 0.565 $\pm$ 0.012 & 0.968 $\pm$0.018 & 0.873 $\pm$0.053 \\
\midrule
\makecell{\textbf{ \ourmethod}}& 0.732 $\pm$0.003 (3) & 0.977 $\pm$0.004 (2) & 0.924 $\pm$0.038 (6)  \\
\bottomrule
\end{tabular}
\caption{Comparing \ourmethod~and other baselines based on the $100^{\text{th}}$  percentile of the solutions (i.e., maximum solution) generated by each method. Each cell shows the mean and standard deviation of the function values found by each method over 4 runs. The individual rank of our method is included next to its reported performance for each benchmark. }
\label{tab:5}
\end{table*}

\begin{table*}[ht]
\centering
\begin{tabular}{lccc} 
\toprule
\textsc{Method} & \textsc{Ant} & \textsc{DKitty} & \textsc{Hopper}  \\
\midrule
\textsc{GA} & 0.130 $\pm$ 0.029& 0.742 $\pm$0.012 & 0.089 $\pm$ 0.07 \\
\textsc{Ens-Mean} & 0.192 $\pm 0.010 $ & 0.791 $\pm$ 0.019 & 0.209 $\pm$ 0.035  \\
\textsc{Ens-Min} & 0.190 $\pm$0.006 & 0.803 $\pm$ 0.005 & 0.166 $\pm$0.052 \\
\textsc{CMA-ES} & -0.049 $\pm$ 0.003 & 0.482 $\pm$ 0.171 & -0.033 $\pm$ 0.006  \\
\textsc{MINS} & 0.614 $\pm$ 0.034 & 0.889 $\pm$ 0.004 & 0.088 $\pm$ 0.170  \\
\textsc{CbAS} & 0.376 $\pm$ 0.023 & 0.757 $\pm$ 0.005 & 0.013 $\pm$ 0.002 \\
\textsc{RoMA} & 0.448 $\pm$ 0.013 & 0.760 $\pm$ 0.028 & 0.370 $\pm$ 0.008  \\
\textsc{BONET} & \textbf{0.620} $\pm$0.003 & \textbf{0.897} $\pm$ 0.0001 & 0.390 $\pm$ 0.0002 \\
\textsc{COMS} & 0.557 $\pm$ 0.015 & 0.879 $\pm$ 0.001& 0.379 $\pm$0.005 \\
\midrule
\makecell{\textbf{\ourmethod}} & 0.611$\pm$ 0.007 (3) & 0.887$\pm$0.003 (3) & \textbf{0.393} $\pm$ 0.005 (1) \\
\midrule
\textsc{Method} & \textsc{SCon} & \textsc{Tf8} & \textsc{Tf10}  \\
\midrule
\textsc{GA} & 0.641 $\pm$ 0.036 & 0.510 $\pm$ 0.055 & 0.794 $\pm$ 0.013 \\
\textsc{Ens-Mean}&  0.644 $\pm$ 0.070 & 0.529 $\pm$ 0.030 & \textbf{0.796} $\pm$ 0.006 \\
\textsc{Ens-Min}& \textbf{0.672} $\pm$ 0.042  & 0.490 $\pm$ 0.052 & 0.794 $\pm$ 0.008\\
\textsc{CMA-ES}&  0.590 $\pm$ 0.012 & 0.592 $\pm$ 0.015 & 0.786 $\pm$ 0.009 \\
\textsc{MINS}&  0.414 $\pm$ 0.011 & 0.420 $\pm$ 0.009 & 0.465 $\pm$ 0.016 \\
\textsc{CbAS}& 0.099 $\pm$ 0.008 & 0.442  $\pm$ 0.038& 0.613 $\pm$ 0.012 \\
\textsc{RoMA}& 0.420 $\pm$ 0.030 & 0.560 $\pm$ 0.104 & 0.780 $\pm$ 0.400 \\
\textsc{BONET}& 0.470 $\pm$ 0.004& 0.505 $\pm$ 0.004& 0.465 $\pm$ 0.002 \\
\textsc{COMS}& 0.414 $\pm$0.023 & \textbf{0.652} $\pm$ 0.108& 0.606 $\pm$ 0.027\\
\midrule
\makecell{\textbf{\ourmethod}} & 0.439 $\pm$ 0.016(6) & 0.594 $\pm$ 0.015(2) & 0.720 $\pm$0.015 (6) \\
\bottomrule
\end{tabular}
\caption{Comparing \ourmethod~and baselines based on $50^{\text{th}}$  percentile of the solutions (i.e., median solution) generated by each method. Each cell shows the mean and standard deviation of the function values found by each method over 4 runs. The individual rank of our method is included next to its reported performance for each benchmark. }
\label{tab:6}
\end{table*}
\newpage
\section{Rank Distribution Plots}
\label{app:rankdist}
To further illustrate the reliability of \ourmethod, this section visualizes the entire rank distribution of the tested algorithm across different percentile level (i.e., 25, 50, 75 and 100). Overall, we observe that \ourmethod~(colored in red) consistently achieves lower mean and standard deviation of performance across all datasets at every percentile level, as compared to that of other baselines. This observation corroborates previous results presented in the main text, and confirms our hypothesis regarding the robustness of \ourmethod.
\begin{figure*}[b]
\centering
\begin{tabular}{cc}
\includegraphics[width=0.47\textwidth]{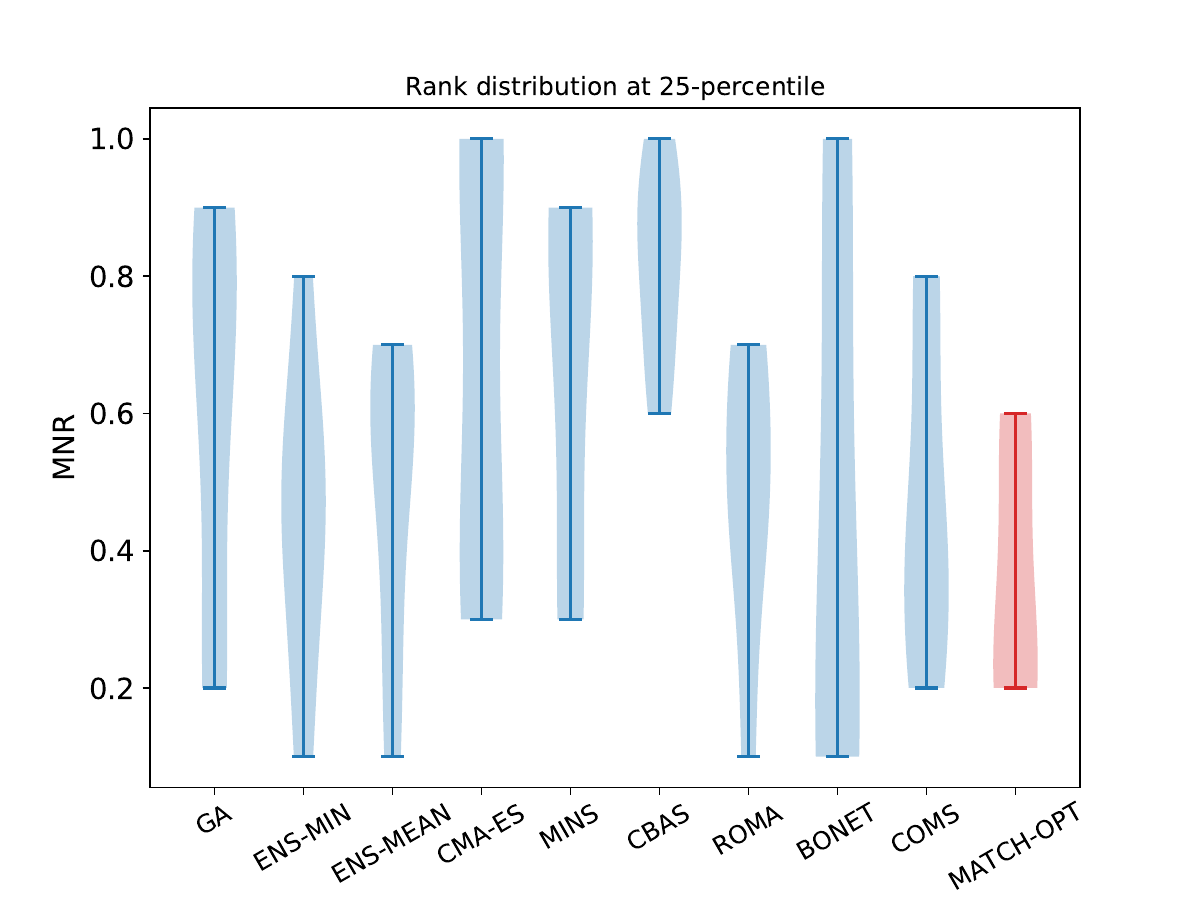} & \includegraphics[width=0.47\textwidth]{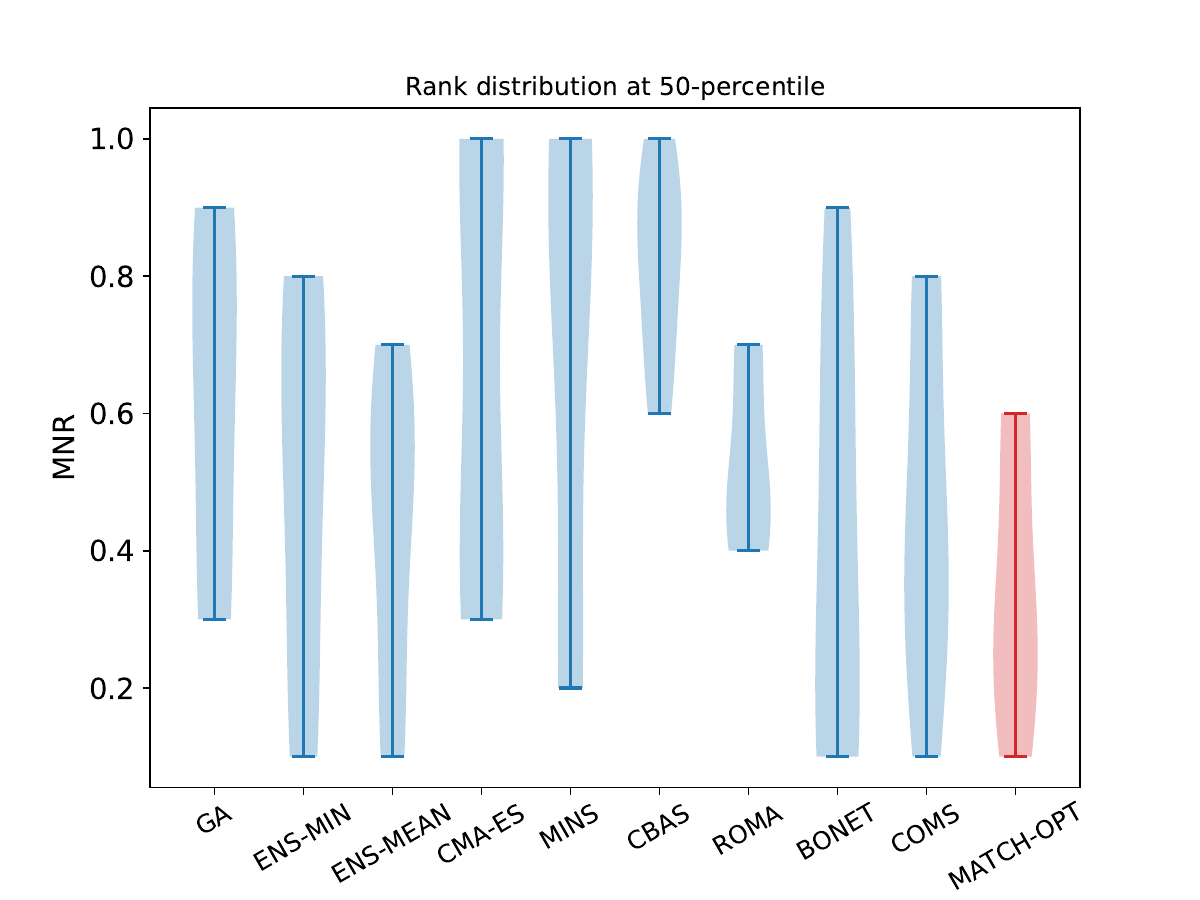} \\
(a) Mean Rank (25th percentile) & (b) Mean Rank (50th percentile) \\
\includegraphics[width=0.47\textwidth]{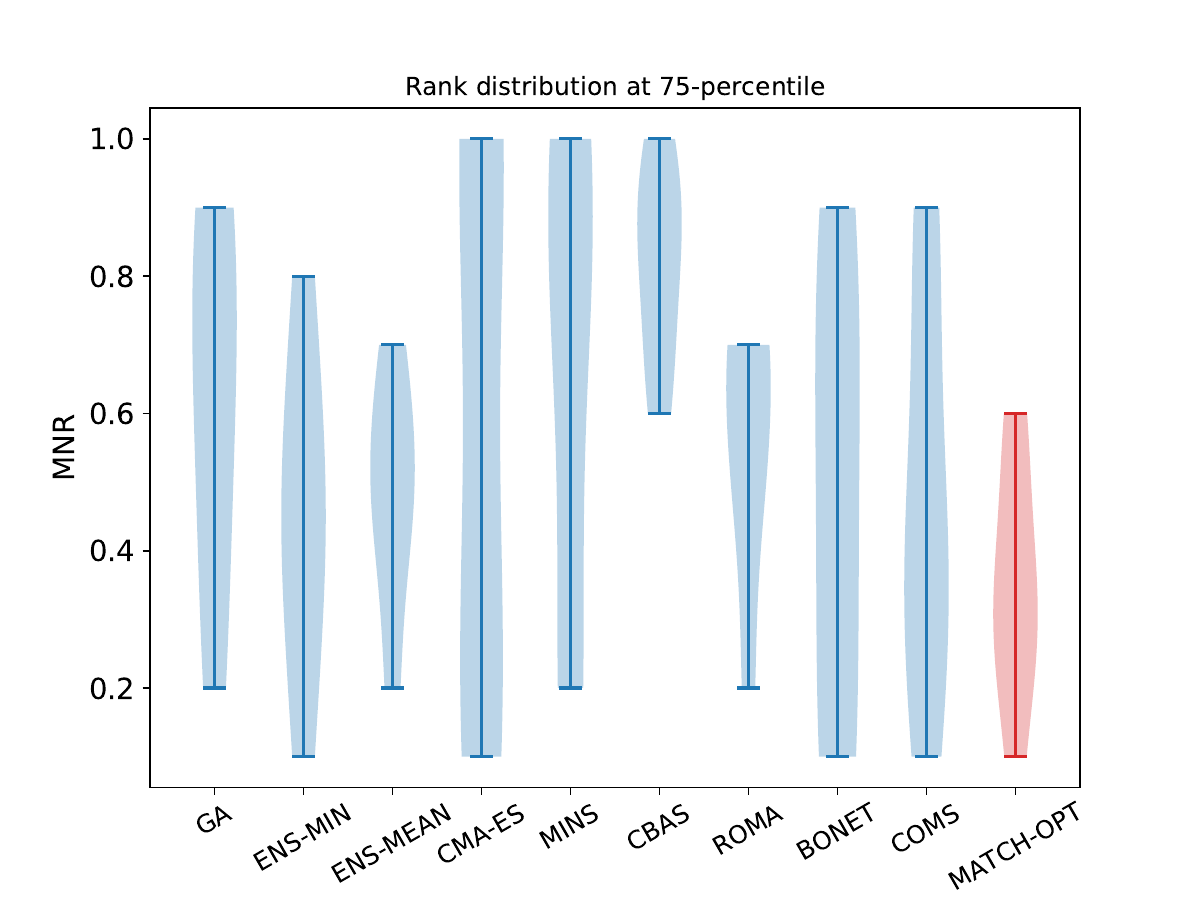} & \includegraphics[width=0.47\textwidth]{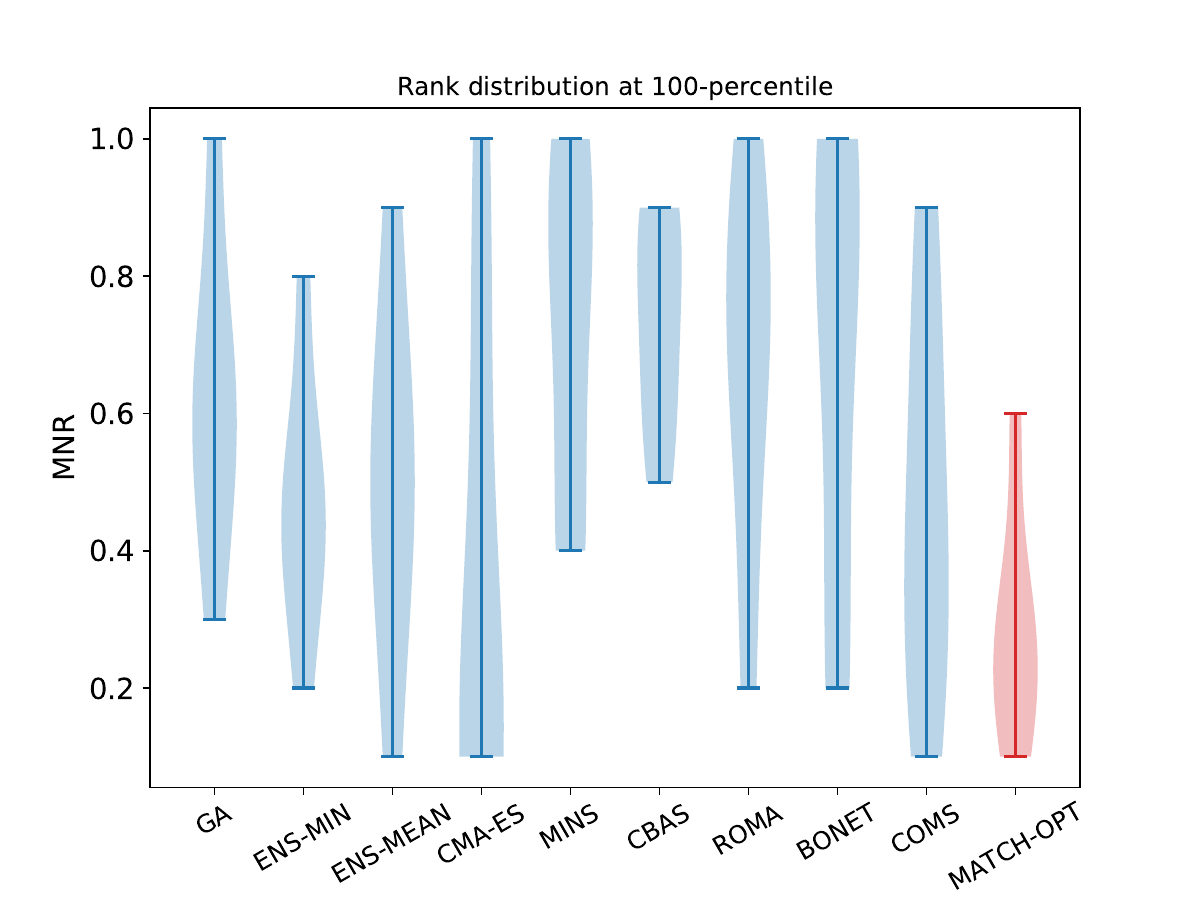}\\
(a) Mean Rank (75th percentile) & (b) Mean Rank (100th percentile)
\end{tabular}
\caption{Plots of distributions of mean normalized rank (\textsc{MNR}) of the tested algorithms across all tasks at the (a) $25$-th, (b) $50$-th, (c) $75$-th, and (d) $100$-th performance percentile levels.} 
\label{fig:rank_violin_plots}
\end{figure*}

\section{Additional Experiments}
\label{app:more_exp}
In addition to the results reported in the main text, we have also compared \ourmethod with three additional baselines, which include DDOM~\cite{krishnamoorthy2023diffusion}, BO-qEI~\cite{wilson2017reparameterization} and BDI~\cite{zhang2022bidirectional}. The results are reported in Table~\ref{tab:100-last} and Table~\ref{tab:50-last} below.

\begin{table*}[hb]
\centering
\begin{tabular}{llllllr} 
\toprule
\hspace{-2mm}\textsc{Method} & \textsc{Ant} & \textsc{DKitty} & \textsc{Hopper} & \textsc{SCon} & \textsc{Tf8} & \textsc{Tf10} \\
\midrule
\hspace{-2mm}\textbf {\ourmethod} & 0.931 & 0.957 & 1.572 & 0.732  & 0.977 & 0.924\\
\hspace{-2mm}\textbf {DDOM} & 0.768 & 0.911 & -0.261 & 0.570  & 0.674 & 0.538\\
\hspace{-2mm}\textbf {BDI} & 0.967 & 0.940 & 1.706 & 0.735  & 0.973 & OOM\\
\hspace{-2mm}\textbf {BO-qEI} & 0.812 & 0.896 & 0.528 & 0.576  & 0.607 & 0.864\\
\bottomrule
\end{tabular}
\caption{Performance comparison between versions of \ourmethod~with DDOM, BO-qEI and BDI at the $100^{\text{th}}$ performance percentile (i.e., maximum solution). OOM indicates that the method runs out of memory} 
\label{tab:100-last}
\end{table*}

\begin{table*}[hb]
\centering
\begin{tabular}{llllllr} 
\toprule
\hspace{-2mm}\textsc{Method} & \textsc{Ant} & \textsc{DKitty} & \textsc{Hopper} & \textsc{SCon} & \textsc{Tf8} & \textsc{Tf10} \\
\midrule
\hspace{-2mm}\textbf {\ourmethod} & 0.611 & 0.887 & 0.393 & 0.439  & 0.594 & 0.720\\
\hspace{-2mm}\textbf {DDOM} & 0.554 & 0.868 & -0.570 & 0.390  & 0.418 & 0.461\\
\hspace{-2mm}\textbf {BDI} & 0.583 & 0.870 & 0.400 & 0.480  & 0.595 & OOM\\
\hspace{-2mm}\textbf {BO-qEI} & 0.568 & 0.883 & 0.360 & 0.490  & 0.439 & 0.557\\
\bottomrule
\end{tabular}
\caption{Performance comparison between versions of \ourmethod~with DDOM, BO-qEI and BDI at the $50^{\text{th}}$ performance percentile (i.e., maximum solution). OOM indicates that the method runs out of memory} 
\label{tab:50-last}
\end{table*}

In both the 50-th and 100-th percentile settings, it appears MATCH-OPT outperforms DDOM in all tasks. Furthermore, the results also show that MATCH-OPT performs the best in 6 out of 12 cases (across both the 100-th and 50-th percentile settings) while BO-qEI only performs best in 1 out of 12 cases. BDI performs best in 5 out of 12 cases, runs out of memory in 2 out of 12 cases. Overall, MATCH-OPT appears to perform more stable than BDI and is marginally better than BDI. It is also more memory-efficient than BDI as it does run successfully in all cases, while BDI runs out of memory in 2 cases. MATCH-OPT also outperforms BO-qEI significantly in 11 out of 12 cases. 

\section{Running Time}
\label{app:running_time}
We also report the running time achieved by all tested algorithms in Table~\ref{tab:running_time}.

\begin{table*}[hb]
\centering
\begin{tabular}{llllllr} 
\toprule
\hspace{-2mm} & \textsc{OURS} & \textsc{BO-qEI} & \textsc{CMA-ES} & \textsc{ROMA} & \textsc{MINS} & \textsc{CBAS} \\
\midrule
\hspace{-2mm} \textsc{Time} & 4785 & 111 & 3804 & 489 & 359 & 189\\
\bottomrule
\end{tabular}

\begin{tabular}{lllllr} 
\hspace{-2mm} & \textsc{BONET} & \textsc{GA} & \textsc{ENS-MEAN} & \textsc{ENS-MIN} & \textsc{DDOM}\\
\midrule
\hspace{-2mm} \textsc{Time} & 614 & 45 & 179 & 179 & 2658 \\
\bottomrule
\end{tabular}
\caption{Total running time (in seconds) of all tested baselines.} 
\label{tab:running_time}
\end{table*}

All reported running times are in seconds. Our algorithm incurs more time than other baselines but its total running time is still affordable in the offline setting: 4785s = 1.32hr. We do, however, want to remark that such complexity comparison is only tangential to our main contribution. Our main focus is on building optimizer with better and more stable performance overall, even at an affordable increase of running time. Furthermore, we want to point out that as some of the baselines (such as BONET) use an entirely different model which has a different number of parameters than ours, the reported running times here might not be comparable on the same compute platform.
The computations were performed on a Ubuntu machine with a 3.73GHz AMD EPYC 7313 16-Core Processor (32 cores, 251 GB RAM) and two NVIDIA RTX A6000 GPUs. Each has 48 GB RAM.
\section{Limitation}
One potential limitation of our approach in comparison to other baselines is that our gradient match algorithm learns from pairs of data points. Thus, the total number of training pairs it needs to consume grows quadratically in the number of offline data points. For example, an offline dataset with $N$ examples will result in a set of $O(N^2)$ training pairs for our algorithm, which increases the training time quadratically. However, an intuition here is that training pairs are not equally informative and, in our experiments, it suffices to get competitive performance by just focusing on pairs of data along the sampled trajectories with monotonically increasing objective function values. This allows us to keep training cost linearly with respect to $N$.

On the other hand, while it is true that none of the existing baselines (including our algorithm) outperform others on all tasks, we believe that at least on these benchmark datasets, our algorithm tends to perform most stably across all tasks, as measured by the mean averaged rank reported in each of our performance tables. This is a single metric that is computed based on the performance of all baselines across all tasks. The end-user can make a judgment based on such metrics. In practice, by looking at how existing baselines perform overall on a set of benchmark tasks that are similar to a target task, one can decide empirically which baseline is most likely to be best for the target task.

\end{document}